\documentclass{article}

\usepackage[preprint]{neurips_2026}

\usepackage[utf8]{inputenc} 
\usepackage[T1]{fontenc}    

\usepackage{url}            
\usepackage{booktabs}       
\usepackage{amsfonts}       
\usepackage{nicefrac}       
\usepackage{microtype}      
\usepackage{xcolor}         

\definecolor{citecolor}{HTML}{0071BC}
\definecolor{linkcolor}{HTML}{ED1C24}
\usepackage[pagebackref=false,breaklinks=true,colorlinks,citecolor=citecolor, linkcolor=linkcolor]{hyperref}

\usepackage{multirow}

\usepackage{multicol}
\usepackage{enumitem}

\usepackage{colortbl}

\definecolor{deepgreen}{rgb}{0.0, 0.5, 0.0}

\usepackage[most]{tcolorbox}
\usepackage{caption}
\usepackage{xcolor}
\usepackage{graphicx}
\usepackage{amsmath}
\usepackage[utf8]{inputenc}

\newtcolorbox[list inside=prompt,auto counter,number within=section]{prompt}[1][]{
    colbacktitle=black!60,
    coltitle=white,
    fontupper=\footnotesize,
    boxsep=5pt,
    enhanced,
    left=0pt,
    right=0pt,
    top=0pt,
    bottom=0pt,
    boxrule=1pt,
    breakable,
    #1
}

\usepackage{makecell}

\usepackage{float}
\usepackage{xspace}
\tcbset{
  aibox/.style={
    width=\textwidth,
    top=10pt,
    colback=white,
    colframe=black,
    colbacktitle=black,
    enhanced,
    center,
    attach boxed title to top left={yshift=-0.1in,xshift=0.15in},
    boxed title style={boxrule=0pt,colframe=white,},
  }
}
\newtcolorbox{AIbox}[2][]{aibox,title=#2,#1}

\definecolor{colorA}{rgb}{0.831, 0.831, 0.902} 
\definecolor{colorB}{rgb}{0.839, 0.953, 0.988} 
\definecolor{colorC}{rgb}{0.847, 0.902, 0.878} 
\definecolor{colorD}{rgb}{0.945, 0.816, 0.780} 
\definecolor{colorE}{rgb}{1.000, 0.965, 0.867} 
\definecolor{customblue}{rgb}{0.1216, 0.4667, 0.7059}
\definecolor{customgreen}{rgb}{0.1725, 0.6275, 0.1725}
\definecolor{customred}{rgb}{0.8392, 0.1529, 0.1569}

\definecolor{colorA}{rgb}{0.831, 0.831, 0.902} 
\definecolor{colorB}{rgb}{0.839, 0.953, 0.988} 
\definecolor{colorC}{rgb}{0.847, 0.902, 0.878} 
\definecolor{colorD}{rgb}{0.945, 0.816, 0.780} 
\definecolor{colorE}{rgb}{1.000, 0.965, 0.867} 
\definecolor{customblue}{rgb}{0.1216, 0.4667, 0.7059}
\definecolor{customgreen}{rgb}{0.1725, 0.6275, 0.1725}
\definecolor{customred}{rgb}{0.8392, 0.1529, 0.1569}

\definecolor{sim_bg_blue}{RGB}{230, 240, 255}
\definecolor{sim_bg_orange}{RGB}{255, 245, 230}
\definecolor{sim_bg_green}{RGB}{235, 250, 235}
\definecolor{sim_bg_gray}{RGB}{230, 230, 230}
\definecolor{sim_text_blue}{RGB}{0, 110, 220}
\definecolor{sim_text_orange}{RGB}{255, 140, 0}
\definecolor{sim_text_green}{RGB}{0, 150, 80}

\tcbuselibrary{listings,breakable}
\usepackage{listings}

\usepackage{textcomp}

\usepackage{wrapfig}

\title{
Agentifying \textit{Patient Dynamics} within LLMs
\\ through  Interacting with Clinical World Model
}

\author{%
    Minghao Wu$^{1\dagger}$, Yuting Yan$^{1\dagger}$, Zhenyang Cai$^{1\dagger}$, Ke Ji$^{1}$, Chuangsen Fang$^{2}$,\\
    \textbf{Ziying Sheng$^{1}$, Xidong Wang$^{1}$, Rongsheng Wang$^{1}$, Hejia Zhang$^{1}$,}\\
    \textbf{Shuang Li$^{1}$, Benyou Wang$^{1*}$, Hongyuan Zha$^{1*}$} \\
    \textsuperscript{\rm 1} The Chinese University of Hong Kong, Shenzhen  \textsuperscript{\rm 2} Beijing Hospital\\
    \texttt{\{wangbenyou, zhahy\}@cuhk.edu.cn} \\
    \url{https://github.com/FreedomIntelligence/SepsisAgent}
}

\begin{document}
\maketitle

\renewcommand{\thefootnote}{\fnsymbol{footnote}}
\footnotetext[2]{Equal Contribution. $^*$Corresponding author.}
\renewcommand{\thefootnote}{\arabic{footnote}}

\begin{abstract}
Sepsis management in the ICU requires sequential treatment decisions under rapidly evolving patient physiology. Although large language models (LLMs) encode broad clinical knowledge and can reason over guidelines, they are not inherently grounded in action-conditioned patient dynamics. We introduce \textbf{SepsisAgent}, a world model-augmented LLM agent for sepsis treatment recommendation. SepsisAgent uses a learned Clinical World Model to simulate patient responses under candidate fluid--vasopressor interventions, and follows a propose--simulate--refine workflow before committing to a prescription. We first show that world-model access alone yields inconsistent LLM decision performance, motivating agent-specific training. We then train SepsisAgent through a three-stage curriculum: patient-dynamics supervised fine-tuning, propose--simulate--refine behavior cloning, and world-model-based agentic reinforcement learning. On MIMIC-IV sepsis trajectories, SepsisAgent outperforms all traditional RL and LLM-based baselines in off-policy value while achieving the best safety profile under guideline adherence and unsafe-action metrics. Further analysis shows that repeated interaction with the Clinical World Model enables the agent to learn regularities in patient evolution, which remain useful even when simulator access is removed.
\end{abstract}


\section{Introduction}
\label{sec:intro}

Sepsis remains a leading cause of mortality in Intensive Care Units (ICUs), presenting a major challenge in critical care medicine~\cite{singer2016third,rudd2020global,rhee2017incidence}. Effective management requires clinicians to titrate intravenous fluids and vasopressors over time to restore perfusion while avoiding downstream organ injury. This makes sepsis treatment a high-stakes \textit{sequential decision-making} problem: actions that improve short-term hemodynamics may still worsen long-term outcomes, for example when aggressive fluid resuscitation restores blood pressure but increases the risk of pulmonary edema or renal failure~\cite{dobson2024revolution,meyhoff2022restriction,douglas2020fluid}. Clinicians must make these decisions under substantial cognitive load, integrating high-dimensional physiological streams~\cite{helman2022engaging} while accounting for patient-specific heterogeneity~\cite{komorowski2018artificial,seymour2019derivation}. This motivates decision-support systems that combine clinical knowledge with patient dynamics~\cite{boussina2024impact}.

\begin{figure}[ht!]
    \centering
    \includegraphics[width=1.0\linewidth]{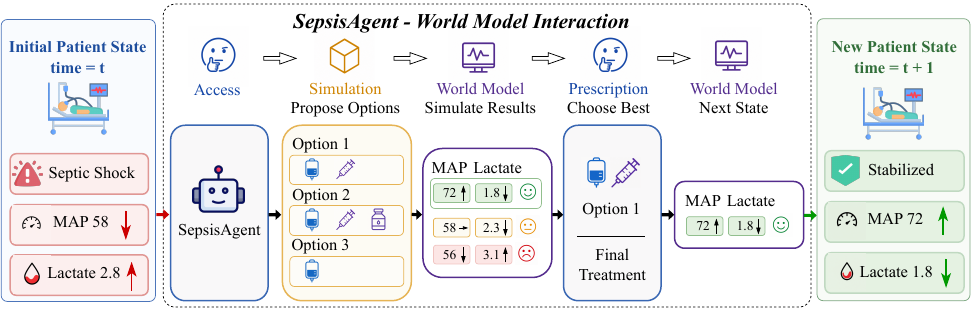}
    \vspace{-0.1 in}
    \caption{Illustration of SepsisAgent's propose--simulate--refine workflow. Given the current patient state, the agent proposes candidate fluid--vasopressor actions, queries the World Model for predicted patient responses, and commits to a final treatment action based on the simulated trajectories.}
    \label{Fig:intro-interaction}
    \vspace{-0.2 in}
\end{figure}

Large Language Models (LLMs) offer a promising interface for clinical decision support because they can interpret heterogeneous clinical context, reason over medical guidelines, and provide natural-language rationales~\cite{xu2025lingshu,sellergren2025medgemma,chen2024huatuogpt}. However, LLMs are not inherently grounded in patient dynamics. They may know that vasopressors increase blood pressure in general, but still fail to estimate how a specific patient will respond to a specific dose over the next few hours~\cite{wornow2023ehrshot}. This limitation is especially problematic in sepsis, where the quality of a treatment decision depends not only on whether it is guideline-consistent, but also on how it changes the patient's future trajectory.

A natural way to address this limitation is to augment the LLM with a predictive model of patient dynamics. We define a \textbf{Clinical World Model} as a learned, action-conditioned approximation of patient evolution. Given a patient state and a candidate treatment action, the world model predicts possible physiological responses and downstream outcome signals. This gives the LLM a mechanism to compare counterfactual treatment options before committing to a prescription. Yet world-model access alone is insufficient: predicted trajectories are approximate, and a generic LLM may over-trust short-term simulated improvements or misinterpret noisy feedback. Therefore, the key challenge is not only to build a simulator, but to train an agent that can use simulated patient responses correctly.

To this end, we introduce \textbf{SepsisAgent}, a world model-augmented LLM agent for sepsis treatment recommendation. SepsisAgent follows a propose--simulate--refine workflow: it proposes candidate fluid--vasopressor interventions, queries the Clinical World Model for simulated patient responses, and refines its final prescription using both simulated feedback and clinical priors. We further train the agent through a three-stage curriculum. First, supervised fine-tuning teaches patient-dynamics prediction and guideline-aware one-step reasoning. Second, behavior cloning teaches multi-round interaction with the world model. Third, world-model-based agentic reinforcement learning optimizes long-horizon treatment strategies through simulated state--action--outcome feedback.

Our contributions are summarized as follows:
\begin{itemize}[leftmargin=*,itemsep=0pt,topsep=0pt,parsep=0pt]
    \item \textbf{A world model-augmented LLM agent for sepsis treatment.}
    We propose \textbf{SepsisAgent}, an LLM-based treatment agent that collaborates with a Clinical World Model to compare candidate fluid--vasopressor interventions before prescription. SepsisAgent outperforms all traditional RL and LLM-based baselines on MIMIC-IV sepsis trajectories.

    \item \textbf{A world-model-based multi-stage training paradigm.}
    We show that merely giving LLMs access to a world model is insufficient. We therefore train SepsisAgent through a multi-stage agentic learning pipeline based on the Clinical World Model, improving its ability to interpret simulated patient responses and optimize treatment policies.

    \item \textbf{Internalizing patient dynamics within LLMs.}
    Our experiments show that world-model-based agentic reinforcement learning enhances the LLM's intrinsic ability to predict patient dynamics in sepsis, rather than merely fitting the reward signal. SepsisAgent improves prediction of in-hospital mortality and vasopressor requirement even without simulator access.
\end{itemize}


\section{Towards World Model Augmented Sepsis Agent}

\label{sec:cwm}

\subsection{Problem Definition}

AI agents are increasingly used to support healthcare decision-making. In this work, we focus on sepsis treatment, a challenging ICU task characterized by rapidly worsening organ dysfunction caused by infection~\cite{singer2016third}. Following prior studies that formulate sepsis treatment as a discrete Markov Decision Process (MDP)~\cite{komorowski2018artificial,raghu2017deep}, we model this task as a sequential decision-making problem. At each decision step \(t\), the agent observes the current patient state \(s_t \in \mathcal{S}\), represented by demographic, physiological, laboratory, and treatment-history variables, and selects a treatment action \(a_t \in \mathcal{A}\).

Following prior sepsis RL formulations~\cite{wu2023value,kalimouttou2025optimal}, our action space covers two key controllable hemodynamic interventions: intravenous fluid administration and vasopressor use, and discretizes these interventions by dosage levels into a \(5 \times 5\) grid:
\[
\mathcal{A}=\mathcal{A}^{\mathrm{fluid}} \times \mathcal{A}^{\mathrm{vaso}}.
\]
The agent aims to recommend treatment actions that stabilize short-term patient physiology while improving long-term clinical outcomes.

\subsection{Motivation to Introduce Clinical World Model}

Most prior methods learn treatment policies directly from retrospective EHR trajectories, mapping the current patient state to a treatment action~\cite{liu2020reinforcement}. Such policies mainly learn associations between observed states and historical clinician actions. They are therefore vulnerable to historical behavior bias and have limited ability to reason about alternative treatment paths that are poorly covered in observational data~\cite{xu2025meddreamer}. However, sepsis treatment is not only about selecting an action for the current state; it also requires estimating how the patient may evolve under different interventions~\cite{raghu2018model}.

This motivates an explicit model of patient response. In model-based reinforcement learning, a world model approximates environment dynamics and supports planning by predicting future states under candidate actions~\cite{ha2018world,lecun2022path}. In this work, \textbf{we define a Clinical World Model as a learned, action-conditioned approximation of patient dynamics}:
\[
W_\theta: (s_t, a_t) \mapsto p_\theta(s_{t+1}, o_t \mid s_t, a_t),
\]
where \(s_t\) is the current patient state representation, \(a_t\) is a candidate treatment action, and \(o_t\) denotes downstream clinical outcomes. For sepsis treatment, this model acts as an approximate simulator that predicts the physiological consequences of fluid--vasopressor interventions, enabling the agent to compare candidate actions before making a prescription.

\subsection{Solution: World Model-Augmented Sepsis Agent}

LLMs and Clinical World Models address different parts of the treatment-decision problem. LLMs provide the semantic layer: they encode clinical prior knowledge, interpret heterogeneous patient context, reason over guidelines, and generate interpretable rationales for treatment decisions~\cite{singhal2023large,singhal2025toward,tu2025towards}. However, they do not directly model how a specific patient state will evolve after a specific intervention~\cite{steinberg2023motor, wornow2024context}. Clinical World Models provide the dynamics layer: they predict action-conditioned patient responses under candidate treatments, but lack the clinical semantic reasoning and guideline-aware judgment of LLMs.

To combine these strengths, we propose \textbf{SepsisAgent}, a world model-augmented LLM agent for sepsis treatment recommendation. The LLM backbone acts as the decision-making policy, while the Clinical World Model acts as an agent component that collaborates with the LLM by predicting patient responses under candidate interventions. Moreover, the world model provides counterfactual exploration trajectories, which are used to further train SepsisAgent to understand patient dynamics and optimize treatment strategies.

At inference time, SepsisAgent follows a propose--simulate--refine workflow. Given the current patient state \(s_t\), the LLM proposes a candidate action set
\[
\mathcal{C}_t = \{a_t^{(1)}, \dots, a_t^{(M)}\}.
\]
For each candidate action, the Clinical World Model estimates its possible physiological consequence:
\[
\hat{y}_t^{(i)} \sim W_\theta(s_t, a_t^{(i)}),
\]
where \(\hat{y}_t^{(i)}\) summarizes the predicted patient response, including future physiological state and outcome-related signals. The LLM then compares these action--response pairs with its clinical priors and commits to the final treatment:
\[
a_t = \pi_\phi\left(s_t, \{(a_t^{(i)}, \hat{y}_t^{(i)})\}_{i=1}^{M}\right).
\]

Building on this formulation, \textbf{Section~3 instantiates the Clinical World Model and evaluates its benefit for SepsisAgent}. \textbf{Section~4 then describes how SepsisAgent explores and learns within this world model}.

\section{Training World Model to Augment Sepsis Agent}

This section builds the Clinical World Model that serves as both the inference-time simulator and the training environment for SepsisAgent. We first construct discrete sepsis trajectories from MIMIC-IV, then instantiate an action-conditioned predictive model of patient dynamics, and finally \textbf{evaluate whether this model provides useful feedback for treatment decision-making.}


We extract 20,092 ICU stays from MIMIC-IV~\cite{johnson2023mimic} using Sepsis-3 criteria~\cite{singer2016third}, covering the window from 24h before to 48h after sepsis onset within each ICU stay. Following AI Clinician~\cite{komorowski2018artificial}, data are aggregated into 4-hour steps and split into training, validation, and test sets (7:2:1). The state space consists of 42 clinical variables (Table~\ref{tab:state_variables}), and the action space is a \(5 \times 5\) discrete grid based on dosage percentiles of intravenous fluids and vasopressors. Further details are provided in Appendix~\ref{app:data_details}.

\subsection{Clinical World Model Instantiation}

\begin{figure}[ht!]
    \vspace{-4mm}
    \centering
    \includegraphics[width=0.8\linewidth]{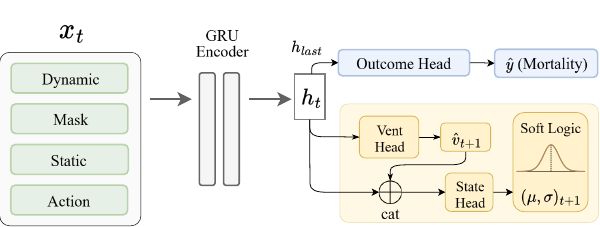}
    \caption{Architecture of the Clinical World Model.}
    \label{Fig::GRU}
\end{figure}

Following MedDreamer~\cite{xu2025meddreamer}, we use a GRU encoder to encode and update latent representations of sepsis patient trajectories (shown in Figure~\ref{Fig::GRU}). Specifically, given the observed trajectory history
\[
\tau_{\leq t}=\{(s_1,a_1),\ldots,(s_{t-1},a_{t-1}),s_t\},
\]
a two-layer GRU encoder produces a history-aware patient representation:
\[
h_t = \mathrm{GRU}_\psi(\tau_{\leq t}).
\]
We adopt this lightweight architecture for proof-of-concept simplicity; as discussed in Appendix~\ref{app:wm_arch_comparison}, increasing model complexity or varying architecture size did not yield significant performance gains.

The World Model is instantiated with two prediction heads. First, the state transition head predicts the next physiological state conditioned on the encoded patient history and candidate treatment action:
\[
p_\theta(s_{t+1}\mid \tau_{\leq t},a_t)
=
\mathcal{N}\left(
\mu_\theta(h_t,a_t),
\Sigma_\theta(h_t,a_t)
\right).
\]
We use the predicted mean \(\hat{s}_{t+1}=\mu_\theta(h_t,a_t)\) as the simulated next-state response returned to the agent. To improve clinical plausibility, auxiliary heads predict ventilation status and derived severity scores, including SOFA and SIRS, which are used as regularization targets during training.

Second, the outcome head estimates longer-term clinical consequences over a fixed prediction window of length \(K\). Given a trajectory segment
\[
\tau_{t:t+K}=\{(s_t,a_t),\ldots,(s_{t+K},a_{t+K})\},
\]
the outcome model predicts the corresponding clinical outcome:
\[
\hat{o}_{t:t+K}
=
\mathcal{W}^{\mathrm{outcome}}_\phi(\tau_{t:t+K}).
\]
Together, the transition and outcome heads allow the world model to provide both short-horizon physiological responses and longer-horizon risk signals for candidate treatment actions. Detailed training configurations are provided in Appendix~\ref{app:wm_training}.

\newpage

\subsection{Clinical World Model as a Predictive Simulator}

\begin{wraptable}{r}{0.4\textwidth}
\vspace{-12mm}
\centering
\caption{Clinical World Model Evaluation.}
\label{tab:wm_evaluation}
\small
\resizebox{0.4\textwidth}{!}{
\begin{tabular}{llc}
\toprule
\textbf{Model Component} & \textbf{Metric} & \textbf{Value} \\
\midrule
State Transition & MAE & 0.316 \\
 & Ventilation AUC & 0.942 \\
\midrule
Outcome Prediction & AUC-ROC & 0.804 \\
 & AUC-PR & 0.663 \\
\bottomrule
\end{tabular}
}
\vspace{-2mm}
\end{wraptable}

We evaluate the Clinical World Model as a predictive simulator and report compact metrics characterizing state prediction accuracy and outcome discrimination (Table~\ref{tab:wm_evaluation}). The results show that the model captures clinical relevant patient dynamics and provides informative short-horizon feedback for candidate treatment actions.

\subsection{Why World-Model Access Alone Is Insufficient}
\label{sec:performance}

To quantify the benefit of the Clinical World Model, we evaluate three sota LLMs (GPT-4.1-mini, Gemini-3-Flash, and o3) under three settings: (1) \textbf{Vanilla LLM}; (2) \textbf{LLM + World Model}, where the model can query simulated patient responses for candidate actions; and (3) \textbf{LLM + World Model + Clinical Prior}, where the model additionally receives concise clinical priors derived from sepsis guidelines~\cite{evans2021surviving}. For efficiency and reproducibility, we randomly sample 725 episodes from the test set as the evaluation benchmark for LLM-based methods\footnote{For Gemini-3-Flash with world model augmentation, the API cost per episode rollout is approximately \$0.31.}. Table~\ref{tab:wm_performance} reports the results.

\begin{table*}[ht!]
\centering
\caption{World-model access alone yields inconsistent LLM decision performance.}
\label{tab:wm_performance}
\small
\setlength{\tabcolsep}{4pt}
\resizebox{\textwidth}{!}{
\begin{tabular}{l|ccc|c|cc}
\toprule
\textbf{Method}
& \multicolumn{3}{c|}{\textbf{Off-Policy Eval}}
& \textbf{Sepsis Guideline}
& \multicolumn{2}{c}{\textbf{Unsafe Actions (\%)}} \\
& \textbf{DR} ($\uparrow$)
& \textbf{WIS} ($\uparrow$)
& \textbf{WPDIS} ($\uparrow$)
& \textbf{Adherence} (\% $\uparrow$)
& \textbf{Underdosing} ($\downarrow$)
& \textbf{Overdosing} ($\downarrow$) \\
\midrule

\textit{Human Reference} & & & & & & \\
Clinicians (Test Set) & 5.06 & 5.27 & 10.82 & 94.76 & 0.35 & 0.19 \\

\midrule
\textit{GPT-4.1-mini} & 6.13 & 6.59 & 10.82 & 80.59 & 0.66 & 2.18 \\
\quad w/ World Model 
& $7.61$\rlap{$_{\textcolor{customgreen}{+}}$}
& $6.69$\rlap{$_{\textcolor{customgreen}{+}}$}
& $9.03$\rlap{$_{\textcolor{customred}{-}}$}
& $84.05$\rlap{$_{\textcolor{customgreen}{+}}$}
& $0.59$\rlap{$_{\textcolor{customgreen}{+}}$}
& $2.87$\rlap{$_{\textcolor{customred}{-}}$} \\
\quad w/ World Model + Clinical Prior 
& $7.31$\rlap{$_{\textcolor{customgreen}{+}}$}
& $5.21$\rlap{$_{\textcolor{customred}{-}}$}
& $17.09$\rlap{$_{\textcolor{customgreen}{+}}$}
& $94.00$\rlap{$_{\textcolor{customgreen}{+}}$}
& $0.58$\rlap{$_{\textcolor{customgreen}{+}}$}
& $1.60$\rlap{$_{\textcolor{customgreen}{+}}$} \\

\midrule
\textit{Gemini-3-Flash} & 8.17 & 9.09 & 13.98 & 96.43 & 1.19 & 2.58 \\
\quad w/ World Model 
& $4.05$\rlap{$_{\textcolor{customred}{-}}$}
& $10.01$\rlap{$_{\textcolor{customgreen}{+}}$}
& $11.33$\rlap{$_{\textcolor{customred}{-}}$}
& $93.62$\rlap{$_{\textcolor{customred}{-}}$}
& $1.90$\rlap{$_{\textcolor{customred}{-}}$}
& $2.60$\rlap{$_{\textcolor{customred}{-}}$} \\
\quad w/ World Model + Clinical Prior 
& $4.49$\rlap{$_{\textcolor{customred}{-}}$}
& $7.78$\rlap{$_{\textcolor{customred}{-}}$}
& $12.24$\rlap{$_{\textcolor{customred}{-}}$}
& $95.16$\rlap{$_{\textcolor{customred}{-}}$}
& $0.78$\rlap{$_{\textcolor{customgreen}{+}}$}
& $1.16$\rlap{$_{\textcolor{customgreen}{+}}$} \\

\midrule
\textit{o3} & 8.32 & 9.17 & 20.38 & 90.55 & 0.72 & 1.57 \\
\quad w/ World Model 
& $8.78$\rlap{$_{\textcolor{customgreen}{+}}$}
& $9.98$\rlap{$_{\textcolor{customgreen}{+}}$}
& $20.17$\rlap{$_{\textcolor{customred}{-}}$}
& $92.08$\rlap{$_{\textcolor{customgreen}{+}}$}
& $0.64$\rlap{$_{\textcolor{customgreen}{+}}$}
& $1.72$\rlap{$_{\textcolor{customred}{-}}$} \\
\quad w/ World Model + Clinical Prior 
& $9.46$\rlap{$_{\textcolor{customgreen}{+}}$}
& $10.27$\rlap{$_{\textcolor{customgreen}{+}}$}
& $22.95$\rlap{$_{\textcolor{customgreen}{+}}$}
& $96.91$\rlap{$_{\textcolor{customgreen}{+}}$}
& $0.09$\rlap{$_{\textcolor{customgreen}{+}}$}
& $0.24$\rlap{$_{\textcolor{customgreen}{+}}$} \\

\bottomrule
\end{tabular}
}
\begin{minipage}{\linewidth}
\vspace{1mm}
\footnotesize
\textit{Note:} Detailed definitions of policy value, guideline adherence, and safety metrics are provided in Section~\ref{sec:setup}. Green ($_{\textcolor{customgreen}{+}}$) indicates improvement over Vanilla, while Red ($_{\textcolor{customred}{-}}$) indicates degradation.
\end{minipage}
\vspace{-4mm}
\end{table*}

The results show that world model augmentation alone is insufficient: generic LLMs may misinterpret simulated patient responses and choose actions that improve short-term signals but hurt longer-term outcomes. Appendix~\ref{app:failure_modes} provides expert-validated failure-mode analysis for such cases.

Adding clinical priors from sepsis guidelines~\cite{evans2021surviving} improves safety metrics in most cases, but does not guarantee policy-value improvement. For example, Gemini-3-Flash with world model and clinical prior performs worse than its vanilla version across all OPE metrics.

\begin{tcolorbox}[colback=blue!5,colframe=blue!35!black,boxrule=0.6pt,left=1mm,right=1mm,top=1mm,bottom=1mm]
\textbf{Key Finding.} World-model access is not enough; the agent must learn how to use it.
\end{tcolorbox}

This observation shifts the focus from giving an LLM access to a world model to training an agent that can use world-model feedback. In Section~4, we turn simulated patient responses into learning signals: SepsisAgent is trained to understand evolving patient dynamics, reason under sepsis guideline priors, and refine treatment policies through repeated interaction with the Clinical World Model.


\section{ Agentifying Patient Dynamics in Agents  through World Model Interaction }

\label{sec:method}

\subsection{From World Model Feedback to Patient-Dynamics Understanding}

\paragraph{Training goal.}
Section~\ref{sec:performance} shows that the bottleneck is not simulator access, but the agent's ability to interpret patient-dynamics feedback and integrate it with clinical priors. From a clinical decision-making perspective, a useful agent should not simply follow the action with the best one-step simulated response. It should first understand the patient's evolving risk, then reason about guideline-consistent treatment choices, and finally refine decisions through counterfactual patient-response estimates. We therefore use the Clinical World Model not only as an inference-time simulator, but also as a training environment that exposes the LLM to repeated state--action--outcome feedback.

\begin{figure*}[t]
    \centering
    \includegraphics[width=1.0\linewidth]{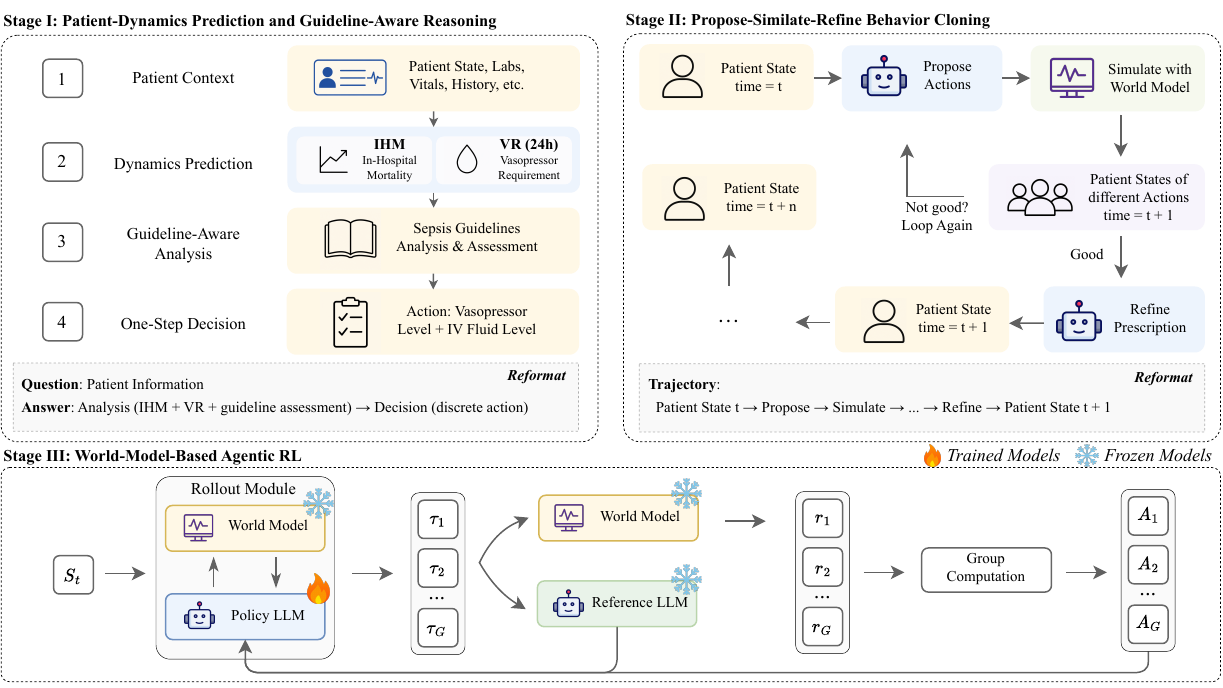}
    \vspace{-0.1 in}
    \caption{Overview of the three-stage training pipeline for SepsisAgent. \textbf{\textit{Stage I}} focuses on guideline-aware patient state understanding, \textbf{\textit{Stage II}} learns agentic interaction with a clinical world model, and \textbf{\textit{Stage III}} applies GRPO, treating the world model as an environment to iteratively predict future patient states and refine treatment policies through rollout interactions.}
    \label{Fig:training-pipeline}
    \vspace{-0.15 in}
\end{figure*}

\paragraph{Curriculum design.}
This motivates a staged curriculum, shown in Figure~\ref{Fig:training-pipeline}. Stage I builds patient-dynamics understanding and guideline-aware treatment reasoning through supervised data that jointly includes IHM prediction, VR prediction, and one-step treatment recommendation. Stage II teaches multi-turn interaction with the world model through imitation learning on synthesized interaction trajectories. Stage III further optimizes the agent through reinforcement learning in the world-model environment, encouraging long-horizon planning rather than greedy short-term stabilization.

\subsection{Stage I: Patient-Dynamics Prediction and Guideline-Aware Reasoning}

\paragraph{Dynamics-aware supervision.}
We construct supervised training data around two prediction tasks defined in the MIMIC-Sepsis benchmark~\cite{huang2025mimic}: in-hospital mortality (IHM) and vasopressor requirement (VR). These tasks are critical for treatment selection, as they capture long-term patient outcome and hemodynamic deterioration.

\paragraph{In-hospital mortality.}
Given the current patient context and treatment information, the model predicts the patient's subsequent in-hospital mortality as a binary outcome. This task encourages the model to attend to the long-term consequences of treatment trajectories rather than only immediate physiological measurements.

\paragraph{Vasopressor requirement.}
The VR task predicts whether the patient will require vasopressor support within the next 24 hours. Based on sepsis guidelines~\cite{evans2021surviving}, VR serves as a clinically meaningful proxy for predicting whether key physiological indicators will cross high-risk clinical thresholds within the next 24 hours.

\paragraph{One-step treatment reasoning.}
We train the model to perform single-step treatment reasoning grounded in patient dynamics and sepsis guideline priors. Each example follows an \textbf{Analysis--Decision} format: in the \textbf{Analysis} phase, the model predicts IHM and VR and evaluates the patient's condition under sepsis guideline priors; in the \textbf{Decision} phase, it recommends the corresponding discrete treatment action. This design makes the two dynamics-prediction tasks part of the treatment reasoning process rather than isolated auxiliary objectives.

\paragraph{Reasoning data construction.}
Following common practice in constructing medical reasoning supervision~\cite{chen2024huatuogpto1,sun2025reasonmed}, we use GPT-4.1 to synthesize chain-of-thought-style rationales from observed clinical facts, expert actions, and sepsis guideline priors. The generated traces are reformatted into structured supervision pairs, enabling the model to learn patient-state analysis and guideline-consistent one-step treatment recommendation.

\subsection{Stage II: Propose--Simulate--Refine Behavior Cloning}

\paragraph{Workflow imitation.}
After Stage I, the model has learned non-agentic patient-dynamics prediction and guideline-aware one-step reasoning. Stage II teaches the model how to use world-model feedback during decision-making: before committing to each treatment action, the agent can query the Clinical World Model for multiple rounds, inspect simulated consequences, and refine its final prescription.

\paragraph{Trajectory synthesis.}
For each expert transition \((s_t, a_t^*)\), we use GPT-4.1 to synthesize a multi-round propose--simulate--refine reasoning trace following the interaction in Section~2.3. The trace starts from the patient state \(s_t\), proposes candidate actions \(\mathcal{C}_t\), obtains simulated responses \(\{\hat{y}_t^{(i)}\}_{i=1}^{M}\) from the Clinical World Model, optionally updates the candidate set based on these responses, and finally selects the expert action \(a_t^*\).

\paragraph{Behavior cloning objective.}
We fine-tune the model to reproduce these structured multi-round traces. This stage provides a cold start for world-model interaction, enabling the agent to learn how simulated patient responses should inform treatment recommendations before reinforcement learning in Stage III.

\subsection{Stage III: World-Model-Based Agentic RL}
\label{sec:agentic_rl}

\paragraph{Long-horizon optimization.}
Behavior cloning provides a cold start for propose--simulate--refine decision-making, but remains constrained by expert demonstrations. In Stage III, we further optimize the policy with Group Relative Policy Optimization (GRPO)~\cite{shao2024deepseekmath}. The agent interacts with the Clinical World Model as a virtual environment, samples multiple reasoning trajectories for each patient state, and learns to select actions that improve long-term patient outcomes rather than merely imitate observed decisions.

\paragraph{World-model rollouts.}
For each initial patient state \(s_0\), the policy \(\pi_\phi\) generates treatment actions through multi-round world-model interaction. The Clinical World Model then rolls out the resulting patient trajectory
\[
\tau = \{(s_0,a_0), (s_1,a_1), \ldots, (s_T,a_T)\},
\]
where each transition is induced by the simulated patient response under the selected treatment action. This allows the agent to receive feedback from complete treatment trajectories, rather than only single-step expert labels.

\paragraph{Composite reward.}
We optimize the policy using a composite reward that combines terminal outcome, intermediate physiological stabilization, and guideline consistency:
\[
R(\tau)=R_{\mathrm{out}}(s_T)+\sum_{t=0}^{T-1} r(s_t,s_{t+1})-\lambda_g \mathcal{P}_g(\tau).
\]
Here, \(R_{\mathrm{out}}\) rewards favorable terminal outcomes such as survival and penalizes mortality, while \(\mathcal{P}_g\) penalizes violations of sepsis guideline constraints.

\paragraph{Intermediate reward.}
Following the intermediate reward design in DDQN for sepsis treatment~\cite{raghu2017deep}, we define the step-wise physiological reward as
\[
r(s_t,s_{t+1})=
C_0\mathbb{I}(\Delta \mathrm{SOFA}_t=0 \land s_{t+1}^{\mathrm{SOFA}}>0)
+C_1\Delta \mathrm{SOFA}_t
+C_2\Delta \mathrm{Lac}_t .
\]
This term encourages improvement in organ dysfunction and lactate dynamics, while the terminal reward encourages long-term survival. Detailed reward coefficients, lactate transformation, clipping strategy, and implementation details are provided in Appendix~\ref{app:training-settings}.


\section{Experiments}
\label{sec:experiments}

\subsection{Experimental Setup}
\label{sec:setup}

\paragraph{Backbone selection.}
We select Qwen3-4B-Instruct~\cite{yang2025qwen3} as the backbone of SepsisAgent. It is sufficiently capable of supporting clinical reasoning and agentic interaction, while remaining efficient for deployment-style evaluation: after training, the average inference time per decision step is 6.1s, well within the minutes-scale decision workflow of ICU sepsis management.

\paragraph{Baselines.}
We compare SepsisAgent with four groups of baselines. \textbf{Clinicians} correspond to the recorded human decisions in the MIMIC-IV test set and serve as the real-world reference policy. \textbf{Traditional RL} baselines include representative sepsis treatment policies, DDQN~\cite{raghu2017deep} and AI Clinician~\cite{komorowski2018artificial}. \textbf{Vanilla LLMs} evaluate whether general-purpose language models can recommend sepsis treatments directly from patient states without world-model feedback. \textbf{World-model-augmented LLMs} use the same Clinical World Model and propose--simulate--refine interaction protocol as SepsisAgent, but do not receive our staged supervised and reinforcement learning training. This comparison isolates the effect of agent training from merely providing world-model access.

\paragraph{Evaluation metrics.}
We evaluate each method on the 725-episode held-out test set from three complementary perspectives. \textbf{Off-policy evaluation} reports DR~\cite{jiang2016doubly}, WIS, and WPDIS~\cite{precup2000eligibility} to estimate policy value from retrospective trajectories without real-world deployment. \textbf{Sepsis Guideline Adherence} measures whether recommended actions satisfy the sepsis guideline constraints used throughout this work~\cite{evans2021surviving}. \textbf{Unsafe Actions} further reports the percentage of extreme underdosing and overdosing actions, following rule-based safety definitions derived from expert clinical practice~\cite{festor2022assuring}. This metric is stricter than guideline adherence, used only as an independent evaluation criterion, and never optimized during training. Detailed rules are provided in Appendix~\ref{app:guideline_safety}.

\begin{table*}[ht!]
\centering
\caption{SepsisAgent achieves the best overall value and safety profile.}
\label{tab:main_results}
\small
\setlength{\tabcolsep}{4pt}
\resizebox{\textwidth}{!}{
\begin{tabular}{l|ccc|c|cc}
\toprule
\textbf{Method} 
& \multicolumn{3}{c|}{\textbf{Off-Policy Eval}} 
& \textbf{Sepsis Guideline} 
& \multicolumn{2}{c}{\textbf{Unsafe Actions (\%)}} \\
& \textbf{DR} ($\uparrow$) 
& \textbf{WIS} ($\uparrow$) 
& \textbf{WPDIS} ($\uparrow$) 
& \textbf{Adherence} (\% $\uparrow$) 
& \textbf{Underdosing} ($\downarrow$) 
& \textbf{Overdosing} ($\downarrow$) \\
\midrule

\rowcolor{colorB!40}
\multicolumn{7}{l}{\textbf{Human Reference}} \\
\midrule
Clinicians (Test Set) & 5.06 & 5.27 & 10.82 & 94.76 & 0.35 & \underline{0.19} \\

\midrule
\rowcolor{colorB!40}
\multicolumn{7}{l}{\textbf{Traditional RL}} \\
\midrule
DDQN~\cite{raghu2017deep} & 8.69 & 6.19 & 15.11 & 82.79 & 0.67 & 1.01 \\
AI Clinician~\cite{komorowski2018artificial} & 5.98 & 9.94 & 9.88 & 87.24& 0.53 & 2.34 \\
WD3QNE~\cite{wu2023value} & 8.72 & \textbf{12.07} & \underline{23.20} & 87.60 & 1.11 & 1.49 \\

\midrule
\rowcolor{colorB!40}
\multicolumn{7}{l}{\textbf{Vanilla LLMs}} \\
\midrule
o3 & 8.32 & 9.17 & 20.38 & 90.55 & 0.72 & 1.57 \\
Gemini-3-Pro~\cite{pichai2025new} & 5.84 & 8.59 & 19.68 & 96.74 & \underline{0.09} & 1.62 \\
Gemini-3-Flash~\cite{pichai2025new} & 8.17 & 9.09 & 13.98 & 96.43 & 1.19 & 2.58 \\
GPT-OSS-120B~\cite{openai2025gptoss120bgptoss20bmodel} & 8.25 & 6.00 & 21.17 & 79.42 & 1.06 & 1.11 \\
GPT4.1-mini~\cite{gpt41} & 6.13 & 6.59 & 10.82 & 80.59 & 0.66 & 2.18 \\
Deepseek-V3.2~\cite{deepseekai2025deepseekv32pushingfrontieropen} & 8.63 & 10.25 & 15.07 & 81.80 & 0.19 & 0.41 \\

\midrule
\rowcolor{colorB!40}
\multicolumn{7}{l}{\textbf{World-Model-Augmented LLMs}} \\
\midrule
o3 + WM & \underline{9.46} & 10.27 & 22.95 & \underline{96.91} & 0.09 & 0.24 \\
Gemini-3-Flash + WM & 4.49 & 7.78 & 12.24 & 95.16 & 0.78 & 1.16 \\
GPT4.1-mini + WM & 7.32 & 5.21 & 17.09 & 94.00& 0.58 & 1.60 \\

\midrule
\rowcolor{colorB!40}
\multicolumn{7}{l}{\textbf{SepsisAgent and Its Backbone}} \\
\midrule
Qwen3-4B-Instruct~\cite{yang2025qwen3} & 7.79 & 7.34 & 18.76 & 78.00 & 0.62 & 2.13 \\
\textbf{SepsisAgent} & \textbf{10.01} & \underline{11.14} & \textbf{23.40} & \textbf{97.95} & \textbf{0.08} & \textbf{0.14} \\

\bottomrule
\end{tabular}
}
\end{table*}

\subsection{Main Results: SepsisAgent Improves Policy Value While Preserving Safety}

Table~\ref{tab:main_results} summarizes the performance of SepsisAgent and all baselines on the 725-episode test set.

\paragraph{SepsisAgent achieves the strongest off-policy value.}
SepsisAgent obtains the best DR and WPDIS scores among all methods, while remaining competitive on WIS. It outperforms both traditional RL and LLM-based policies, and also improves consistently over its base model across all three OPE estimators. This suggests that our staged training pipeline benefits from the LLM's clinical prior knowledge while further improving treatment policy quality.

\paragraph{SepsisAgent achieves the best safety profile.}
SepsisAgent achieves the highest sepsis guideline adherence and the lowest unsafe-action rates, indicating that the agent does not improve estimated policy value by taking unsafe treatment shortcuts. Traditional RL policies tend to produce more extreme overdosing actions. Clinician behavior is more conservative, but can still contain suboptimal underdosing patterns, consistent with prior observations in sepsis treatment practice~\cite{wu2023value,festor2022assuring}.

\paragraph{Training matters beyond world-model access.}
World-model-augmented LLMs improve over many vanilla LLM baselines, but they still fall short of SepsisAgent. This suggests that simulated patient responses alone are not enough. The agent must be trained to interpret dynamics feedback and align it with clinical safety constraints.

\subsection{Ablation Study: RL Drives Policy Improvement}

Table~\ref{tab:world_model_eval} reports the cumulative effect of each training stage. The OPE results show that policy-value improvement mainly comes from the RL stage. SFT and behavior cloning improve some metrics, but do not yield stable gains across DR, WIS, and WPDIS. The safety metrics improve earlier: Stage I and Stage II already teach the model to reason within sepsis guideline and expert-defined safety boundaries. Stage III further improves both safety metrics, suggesting that the Clinical World Model provides an effective RL training environment for optimizing long-horizon treatment value while preserving clinically safe behavior.

\begin{table*}[ht!]
\centering
\caption{Staged training improves policy value, safety, and internalization of patient dynamics.}
\label{tab:world_model_eval}
\small
\setlength{\tabcolsep}{4pt}
\resizebox{\textwidth}{!}{
\begin{tabular}{l|ccc|c|c|cc|cc}
\toprule
\textbf{Method}
& \multicolumn{3}{c|}{\textbf{Off-Policy Eval} ($\uparrow$)}
& \textbf{Sepsis Guideline}
& \textbf{Unsafe}
& \multicolumn{2}{c|}{\textbf{IHM} ($\uparrow$)}
& \multicolumn{2}{c}{\textbf{VR} ($\uparrow$)} \\
& \textbf{DR}
& \textbf{WIS}
& \textbf{WPDIS}
& \textbf{Adherence} (\% $\uparrow$)
& \textbf{Actions (\% $\downarrow$)}
& \textbf{AUROC}
& \textbf{AUPRC}
& \textbf{AUROC}
& \textbf{AUPRC} \\
\midrule

\rowcolor{colorB!40}
\multicolumn{10}{l}{\textbf{Base Model}} \\
\midrule
Qwen3-4B-Instruct
& 7.79 & \underline{7.34} & 18.76 & 78.00 & 2.75
& 65.27 & 45.01 & 70.62 & 61.74 \\

\midrule
\rowcolor{colorB!40}
\multicolumn{10}{l}{\textbf{SepsisAgent Variants}} \\
\midrule
SepsisAgent (Stage I: SFT)
& \underline{9.21} & 7.17 & 19.56 & 88.01 & 1.09
& 67.50 & \underline{50.25} & \underline{76.40} & \underline{65.11} \\
SepsisAgent (Stage I+II: +BC)
& 8.99 & 6.81 & \underline{19.61} & \underline{96.89} & \underline{0.51}
& \textbf{67.55} & 46.63 & 74.56 & 63.70 \\
\textbf{SepsisAgent (Stage I+II+III: +RL)}
& \textbf{10.01} & \textbf{11.14} & \textbf{23.40} & \textbf{97.95} &
\textbf{0.22}& \underline{68.52} & \textbf{53.45} & \textbf{79.96} & \textbf{68.83} \\

\bottomrule
\end{tabular}
}
\vspace{-3mm}
\end{table*}

\subsection{Analysis: Internalizing Patient Dynamics within LLMs}

We test whether SepsisAgent internalizes patient dynamics by evaluating IHM and VR prediction without access to the Clinical World Model. As shown in Table~\ref{tab:world_model_eval}, Stage III achieves the best IHM AUPRC and the best VR AUROC/AUPRC, while maintaining competitive IHM AUROC. This suggests that world-model-based agentic RL improves the LLM's intrinsic ability to predict patient outcomes and future vasopressor needs. The agent therefore does not merely fit the reward signal; repeated interaction with the Clinical World Model helps it learn patient-dynamics regularities that remain useful even when simulator access is removed.

\section{Conclusion}

We presented \textbf{SepsisAgent}, a world model-augmented LLM agent for sepsis treatment recommendation. SepsisAgent uses a Clinical World Model to compare counterfactual treatment responses and refine fluid--vasopressor prescriptions. Our results show that world-model access alone is insufficient for generic LLMs, motivating a staged training pipeline that teaches the agent to interpret simulated patient dynamics and align them with clinical priors. On MIMIC-IV sepsis trajectories, SepsisAgent achieves the strongest off-policy value and the best safety profile among traditional RL and LLM-based baselines. Ablation studies further show that world-model-based RL drives policy-value improvement while preserving safety. Finally, intrinsic prediction results indicate that repeated interaction with the Clinical World Model helps the LLM internalize patient dynamics, improving mortality and vasopressor-requirement prediction even without simulator access. These findings provide a proof of concept for using world-model-based agentic learning to support sequential clinical decision-making in high-acuity settings.

\clearpage
\nocite{langley00}

\section*{Acknowledgments}
This work was supported by Major Frontier Exploration Program (Grant No. C10120250085) from the Shenzhen Medical Academy of Research and Translation (SMART),  Shenzhen Medical Research Fund (B2503005),  the Shenzhen Science and Technology Program (JCYJ20220818103001002), NSFC grant 72495131, Shenzhen Doctoral Startup Funding (RCBS20221008093330065), Tianyuan Fund for Mathematics of National Natural Science Foundation of China (NSFC) (12326608), Shenzhen Science and Technology Program (Shenzhen Key Laboratory Grant No. ZDSYS20230626091302006), the 1+1+1 CUHK-CUHK(SZ)-GDSTC Joint Collaboration Fund, Guangdong Provincial Key Laboratory of Mathematical Foundations for Artificial Intelligence (2023B1212010001),  the International Science and Technology Cooperation Center, Ministry of Science and Technology of China (under grant 2024YFE0203000), and Shenzhen Stability Science Program 2023.

\bibliography{ourbib}
\bibliographystyle{unsrt}

\newpage
\appendix
\onecolumn

\section{Related Work}

\subsection{LLM Agent Using a World Model}
\label{subsec:policy_model}

Sepsis treatment involves complex and patient-specific hemodynamic dynamics, and current practices in high-acuity clinical settings are known to be suboptimal~\cite{nauka2025challenges}. The heterogeneous and temporally evolving nature of patient responses makes it difficult to determine optimal interventions using static clinical guidelines alone~\cite{steinberg2023motor,huang2025mimic}.

To address this challenge, we draw inspiration from the notion of world models in model-based reinforcement learning~\cite{ha2018world,lecun2022path} and instantiate it as a predictive simulator of patient physiology. In our framework, an LLM serves as the decision-making agent and queries the Clinical World Model during action selection. Given candidate treatment strategies, the world model provides approximate simulations of short-term physiological responses in sepsis patients. This predictive information complements the LLM's clinical reasoning and supports more informed treatment decisions.

\subsection{Model-based Agentic Reinforcement Learning}

Model-based reinforcement learning has been widely explored for treatment recommendation~\cite{xu2025meddreamer,raghu2018model,perera2026smart}, as it enables policy optimization through explicit modeling of patient dynamics and simulated rollouts. Compared to model-free approaches~\cite{liu2024reinforcement,estiri2024model,luo2024reinforcement}, model-based methods offer improved sample efficiency and the ability to reason about the consequences of treatment decisions in settings where real-world interaction is not feasible.

Building on this paradigm, we further construct the Clinical World Model as a virtual environment for training the LLM policy. Within this environment, the LLM performs simulated rollouts of treatment decisions and observes the resulting physiological trajectories~\cite{xu2025meddreamer,raghu2018model,killian2017robust}.

Through repeated interaction with the world model, the LLM can move beyond purely supervised imitation by incorporating simulated feedback into its decision process. Rather than treating model predictions as an oracle, this formulation allows the policy to reason over predicted trajectories in conjunction with its clinical priors. Moreover, the model-based setup naturally supports decision-making over multiple time steps, facilitating the consideration of both short-term physiological responses and longer-term clinical objectives.

\section{Impact Statement}

This paper presents research aimed at advancing machine learning methods for sequential decision-making. While the proposed approach is motivated by clinical decision support, it is evaluated solely using retrospective data and simulated environments and is not intended for direct clinical deployment. As with many applications of machine learning in healthcare, there are potential societal and ethical considerations related to model uncertainty, data bias, and the need for appropriate human oversight, which we do not believe require further discussion here.

\section{Limitations}

This study is conducted solely for academic research purposes. The simulation results and models presented in this paper are not intended for real-world clinical decision-making. Any potential clinical use should be conducted only under the supervision of doctors.

\section{Data Sources Details}
\label{app:data_details}

We use data from the Medical Information Mart for Intensive Care IV (MIMIC-IV, v2.2)~\cite{johnson2023mimic}. We identify eligible patients using the Sepsis-3 criteria~\cite{singer2016third}, selecting ICU stays with suspected infection (antibiotics and blood cultures within $\pm 24$ hours) and a SOFA score $\geq 2$. We exclude patients under 18, stays less than 24 hours, and cases with missing mortality data.

\paragraph{Data Preprocessing.}
We align our data extraction and preprocessing pipeline with the AI Clinician study~\cite{komorowski2018artificial}. Given the schema differences between MIMIC-III (used in prior work) and MIMIC-IV, we resolved variable mapping conflicts by referencing the official \texttt{mimic-code} repository\footnote{\url{https://github.com/MIT-LCP/mimic-code}}. Unlike previous works that use irregular time steps or 1-hour intervals, we aggregate data into 4-hour time steps to align with the clinical decision-making rhythm for stable sepsis management.
\begin{itemize}
    \item \textbf{Feature Aggregation Logic:} Data is aggregated into 4-hour time steps starting from ICU admission ($t=0$).
    \begin{itemize}
        \item \textit{Summation:} Total Effective Volume (TEV) and Urine Output are summed within the window. Cross-window fluid infusions are allocated proportionally based on duration.
        \item \textit{Maximum:} Vasopressors (NE-Eq) use the maximum infusion rate within the window. Ventilation status takes the most severe state (Invasive > NIV > HFNC > O2 > None).
        \item \textit{Mean/Worst:} Vital signs (HR, MAP, SpO$_2$) are averaged, while GCS uses the worst (minimum) value to capture neurological decline.
        \item \textit{Last Value:} Laboratory results use the last recorded value (sample-and-hold). SOFA scores are recalculated post-aggregation.
    \end{itemize}
    \item \textbf{Imputation Strategy:} Missing values are handled via forward filling. If no prior value exists (e.g., at the start of an admission), we impute using the population median derived from the training set.
\end{itemize}

\paragraph{Action Space Definition.}
We define the two action dimensions as follows (detailed in Table~\ref{tab:action_variables}):
\begin{enumerate}
    \item \textbf{Norepinephrine Equivalent (NE-Eq):}\cite{brown2013survival} Aggregates multiple vasopressors into a standard norepinephrine scale. Note that Dopamine is excluded from our final action space due to its declining usage in modern sepsis protocols.
    \item \textbf{Total Effective Volume (TEV):}\cite{waechter2014interaction} Aggregates crystalloids and colloids based on their volume expansion effect. Dextrose 5\% is excluded as it functions as free water rather than a volume expander.
\end{enumerate}

\begin{table}[h]
\centering
\caption{Cohort statistics stratified by 90-day mortality.}
\label{tab:cohort_stats}
\small
\setlength{\tabcolsep}{3.5pt}
\resizebox{0.48\textwidth}{!}{
\begin{tabular}{lcccc}
\toprule
\textbf{Group} & \textbf{\% Female} & \textbf{Mean Age} & \textbf{Avg Steps} & \textbf{Population} \\
\midrule
Survivors & 42.8 & 61.9 & 11.7 & 13,446 \\
Non-survivors & 43.0 & 68.0 & 11.5 & 6,646 \\
\bottomrule
\end{tabular}
}
\end{table}

\begin{table*}[ht!]
\centering
\caption{State observation variables ($D=42$) grouped by physiological system. Data is aggregated into 4-hour time steps.}
\label{tab:state_variables}
\small
\resizebox{\textwidth}{!}{
\begin{tabular}{@{}l p{14cm}@{}}
\toprule
\rowcolor{colorB!40} \textbf{Category} & \textbf{Variables} \\
\midrule
\textbf{Demographics (5)} & Age, Gender, Weight, ICU Readmission Status, Elixhauser Comorbidity Index \\
\midrule
\textbf{Vital Signs (8)} & Heart Rate, Mean Arterial Pressure (MAP), Respiratory Rate, SpO$_2$, Temperature, Glasgow Coma Scale (GCS) Total, Shock Index, Urine Output \\
\midrule
\textbf{Metabolic \& Renal (10)} & pH, Lactate, Bicarbonate, Base Excess, Anion Gap, BUN, Creatinine, Sodium, Potassium, Chloride, Glucose \\
\midrule
\textbf{Hematology (7)} & Hemoglobin, Hematocrit, White Blood Cell (WBC) Count, Platelet Count, INR, PT, PTT \\
\midrule
\textbf{Organ Function (6)} & Total Bilirubin, Albumin, ALT, AST, PaO$_2$/FiO$_2$ Ratio, SOFA Score (Total \& Sub-scores) \\
\midrule
\textbf{Respiratory (3)} & Mechanical Ventilation Status, FiO$_2$, PaCO$_2$ \\
\midrule
\textbf{Others (3)} & Total CO$_2$, Calcium (Total/Ionized), Magnesium \\
\bottomrule
\end{tabular}
}
\end{table*}

\begin{table*}[ht!]
\centering
\caption{Action space aggregation and discretization. The agent selects a discrete level (0-4) for both dimensions simultaneously.}
\label{tab:action_variables}
\small
\resizebox{\textwidth}{!}{
\begin{tabular}{@{}l p{5cm} p{8cm}@{}}
\toprule
\rowcolor{colorB!40} \textbf{Action Dimension} & \textbf{Unit} & \textbf{Aggregation Formula} \\
\midrule
\textbf{1. Vasopressors} & \multirow{2}{*}{mcg/kg/min (NE-Eq)} & $\text{NE-Eq} = \text{Norepinephrine} + \text{Epinephrine} + \text{Phenylephrine}/10 $ \\
(Norepinephrine Equivalent) & & $\quad\quad\quad + \text{Dopamine}/100 + \text{Vasopressin} \times 2.5/60$ \\
\midrule
\textbf{2. IV Fluids} & \multirow{2}{*}{mL/4h (TEV)} & $\text{TEV} = \sum_{k} w_k V_k$ (See Appendix Table~\ref{tab:tev_mapping} for full ItemID curation) \\
(Total Effective Volume) & & $\quad\quad\quad + 2 \times (V_{\text{Albumin 5\%}} + 5 \times V_{\text{Albumin 25\%}})$ \\
\bottomrule
\end{tabular}
}
\end{table*}

\begin{table*}[ht!]
\centering
\caption{Detailed curation of Total Effective Volume (TEV) coefficients. We mapped over 20 distinct MIMIC-IV ItemIDs to their physiological expansion coefficients ($w_k$), ranging from 0.25 (hypotonic) to 8.0 (hypertonic).}
\label{tab:tev_mapping}
\resizebox{\textwidth}{!}{
\begin{tabular}{@{}llp{8cm}@{}}
\toprule
\textbf{Coeff ($w_k$)} & \textbf{Fluid Type} & \textbf{MIMIC-IV ItemIDs} \\ \midrule
$0.25\times$ & Saline 0.255\% & 220958 \\
$0.30\times$ & Saline 0.3\% & 220959 \\
$0.50\times$ & NaCl 0.45\%, D5 1/2NS & 225159, 225823, 220965 \\
$1.00\times$ & NaCl 0.9\%, LR, Plasma-Lyte & 225158, 225828, 226372, etc. \\
$2.00\times$ & Albumin 5\%, FFP, Platelets & 220864, 220970, 225168, 225170, 225171, 221000, 221013 \\
$2.75\times$ & Mannitol & 227531 \\
$3.00\times$ & NaCl 3\% & 225161 \\
$5.00\times$ & Albumin 25\% & 220862 \\
$6.66\times$ & Sodium Bicarbonate 8.4\% & 220995, 227533 \\
$8.00\times$ & NaCl 23.4\% & 228341 \\ \bottomrule
\end{tabular}
}
\end{table*}

\section{World Model Training Details}
\label{app:wm_training}

\subsection{Architecture Details}
We implement the Clinical World Model using PyTorch. The framework consists of a shared temporal encoder that extracts joint representations for both state transition modeling and outcome prediction.

\begin{itemize}
    \item \textbf{Shared Backbone:} The backbone is a 2-layer Gated Recurrent Unit (GRU) with a hidden dimension of 128. To prevent overfitting, a dropout rate of 0.2 is applied between the GRU layers. The encoder takes a concatenated input of dynamic features, missingness masks, embedded static features (32-dimensional), and embedded treatment actions (32-dimensional).
    \item \textbf{State Prediction Head:} This head follows a hierarchical structure. It first predicts the ventilation status $\hat{v}_{t+1}$ via a Multi-Layer Perceptron (MLP). The hidden state $h_t$, augmented with the predicted ventilation status, is then projected to the parameters of a Gaussian distribution ($\mu, \sigma$) for each of the dynamic physiological variables.
    \item \textbf{Outcome Head:} The outcome head is an MLP with a hidden layer of size 64 and ReLU activation. It takes the joint temporal representation $h_{last}$ from a 48-hour trajectory window to predict the probability of 90-day mortality.
    \item \textbf{Soft Logic Layer:} To maintain clinical consistency, a differentiable Soft Logic layer reverses the normalization and log-transformations of the predicted states. It computes SOFA and SIRS scores using sigmoid-based soft thresholds with a temperature parameter $\tau = 10.0$.
\end{itemize}

\subsection{Training and Optimization}
The model is trained end-to-end by minimizing a composite multi-task loss function:
\begin{equation}
\mathcal{L} = \mathcal{L}_{\text{NLL}} + \lambda_1 \mathcal{L}_{\text{Outcome}} + \lambda_2 \mathcal{L}_{\text{Reg}} + \lambda_3 \mathcal{L}_{\text{Vent}}
\end{equation}
where $\mathcal{L}_{\text{NLL}}$ is the Gaussian negative log-likelihood of the next state, $\mathcal{L}_{\text{Outcome}}$ is the binary cross-entropy loss for mortality, and $\mathcal{L}_{\text{Reg}}$ is the Smooth-L1 loss for SOFA/SIRS consistency. 

We use the \textbf{AdamW} optimizer with a learning rate of $1e-3$ and a weight decay of $1e-4$. The learning rate is dynamically adjusted using a \texttt{ReduceLROnPlateau} scheduler with a reduction factor of 0.5 and a patience of 3 epochs. Training is conducted with a batch size of 2048 for 50 epochs, utilizing early stopping with a patience of 8 epochs based on validation performance.

\subsection{Hyperparameter Settings}

For the shared clinical world model, we used a two-layer GRU backbone with a hidden dimension of 128 and a dropout rate of 0.2. Static patient features and treatment actions were separately projected into 32-dimensional embeddings before being integrated into the temporal modeling framework. The trajectory window size was set to $K=12$, corresponding to a 48-hour observation window. During training, we used a batch size of 2048 and optimized the model with AdamW using a learning rate of $1 \times 10^{-3}$ and a weight decay of $1 \times 10^{-4}$. The loss weights were set to $\lambda_1=1.0$ for outcome prediction, $\lambda_2=0.01$ for consistency regularization, and $\lambda_3=0.3$ for ventilation prediction. Early stopping was applied with a patience of 8 epochs to prevent overfitting.

\subsection{Performance of Different World Models}
\label{app:wm_arch_comparison}

We compare four candidate architectures for the Clinical World Model: Transformer, LSTM, 2-layer GRU, and 8-layer GRU. Table~\ref{tab:wm_arch_prediction} reports predictive performance of the world models themselves, while Table~\ref{tab:wm_arch_agent_ope} reports the off-policy evaluation results of SepsisAgent trained with each corresponding world model.

\begin{table}[ht!]
\centering
\caption{Predictive performance of different Clinical World Model architectures.}
\label{tab:wm_arch_prediction}
\small
\setlength{\tabcolsep}{5pt}
\resizebox{\textwidth}{!}{
\begin{tabular}{lcccc}
\toprule
\textbf{World Model} & \textbf{State MAE} ($\downarrow$) & \textbf{Ventilation AUC} ($\uparrow$) & \textbf{Outcome AUC-ROC} ($\uparrow$) & \textbf{Outcome AUC-PR} ($\uparrow$) \\
\midrule
Transformer & \textbf{0.292} & \textbf{0.951} & \textbf{0.821} & \textbf{0.686} \\
LSTM & \underline{0.299} & 0.940 & 0.805 & \underline{0.670} \\
GRU-8layers & 0.308 & \underline{0.949} & \underline{0.810} & 0.657 \\
GRU-2layers & 0.316 & 0.942 & 0.804 & 0.663  \\

\bottomrule
\end{tabular}
}
\end{table}

\begin{table}[ht!]
\centering
\caption{Off-policy evaluation of SepsisAgent trained with different Clinical World Models.}
\label{tab:wm_arch_agent_ope}
\small
\setlength{\tabcolsep}{8pt}
\begin{tabular}{lccc}
\toprule
\textbf{Method} & \multicolumn{3}{c}{\textbf{Off-Policy Eval} ($\uparrow$)} \\
\cmidrule(lr){2-4}
 & \textbf{DR} & \textbf{WIS} & \textbf{WPDIS} \\
\midrule
SepsisAgent-Transformer WM & \textbf{12.06} & \textbf{12.10} & 22.10 \\
SepsisAgent-LSTM WM & 9.87 & \underline{10.62} & \textbf{23.50} \\
SepsisAgent-GRU-8layers WM & 10.13 & \underline{11.99} & 22.95 \\
SepsisAgent-GRU-2layers WM & 10.01 & 11.14 & \underline{23.40} \\

\bottomrule
\end{tabular}
\end{table}

Tables~\ref{tab:wm_arch_prediction} and~\ref{tab:wm_arch_agent_ope} show that the Transformer world model achieves the best predictive performance across all metrics, but SepsisAgent trained with it does not show a consistent OPE advantage over recurrent world models. Nevertheless, we observe that more accurate world models tend to bring better DR and WIS improvements, suggesting that future Clinical World Models with higher predictive fidelity may further benefit SepsisAgent. Since the 2-layer GRU world model is already sufficient for SepsisAgent to outperform all baselines in the main experiments, we adopt this simplified architecture for proof-of-concept simplicity.

\section{Details of Sepsis Guidelines and Safety Metrics}
\label{app:guideline_safety}

\paragraph{Sepsis guideline priors.}
We use the following sepsis guideline priors derived from the \textit{Surviving Sepsis Campaign 2021} guidelines~\cite{evans2021surviving} and clinical expert review:
\begin{itemize}[leftmargin=*]
    \item Sepsis and septic shock are medical emergencies, and treatment and resuscitation should begin immediately.
    \item For patients with sepsis-induced hypoperfusion or septic shock, at least 30 mL/kg of IV crystalloid fluid should be given within the first 3 hours of resuscitation.
    \item For adults with septic shock on vasopressors, an initial target Mean Arterial Pressure (MAP) of 65 mmHg is recommended over higher targets.
    \item Vasopressors should be initiated if MAP remains below 65 mmHg after adequate fluid resuscitation.
\end{itemize}

\paragraph{Unsafe action metrics.}
In addition to guideline adherence, we report expert-defined unsafe-action rates following rule-based safety evaluation protocols~\cite{festor2022assuring}. These rules are stricter than the guideline-adherence metric and are used only for evaluation, never as training objectives. We define two unsafe-action types:
\begin{itemize}[leftmargin=*]
    \item \textbf{Extreme underdosing:} the patient is severely hypotensive, but the recommended treatment provides no vasopressor support and no or low IV fluid. Formally, this occurs when
    \[
    \mathrm{MAP} < 55\ \mathrm{mmHg}, \quad a^{\mathrm{vaso}} = 0, \quad a^{\mathrm{fluid}} \leq \mathrm{Low}.
    \]
    \item \textbf{Extreme overdosing:} the patient is hypertensive, but the recommended treatment still gives a high vasopressor dose. Formally, this occurs when
    \[
    \mathrm{MAP} > 95\ \mathrm{mmHg}, \quad a^{\mathrm{vaso}} > \mathrm{High}.
    \]
\end{itemize}

\section{Details of the Three-stage Training pipeline}
\label{app:czy-training}

This section provide a detailed description of the training data construction for \textit{Stage I} and \textit{Stage II}, followed by the GRPO-based agentic reinforcement learning procedure.

\subsection{Data Construction}

\subsubsection{\textit{Stage I} Training Data}
ICU data contain a large amount of numerical state information. To help the LLM better understand and analyze such data in an aligned manner, we introduce this training stage and construct targeted training data accordingly. Specifically, each data instance consists of structured ICU clinical signals $\mathcal{S}$, including multivariate physiological and laboratory measurements over a fixed time window, together with a task-specific instruction. The training data are constructed by prompting GPT-4.1 to perform diverse instruction-driven tasks on the same clinical signals, producing analyses from different perspectives, thereby enhancing the model's ability to understand patient states.

In Stage I, we leverage a strong teacher LLM to generate supervision signals under multiple complementary clinical tasks. Each data instance provides the teacher model with the patient state $\mathcal{S}$, relevant clinical guidelines $g$, and reference physician actions $a^{*}$. Task-specific instructions are designed to prompt the teacher to analyze the same patient state from different clinical perspectives.

Specifically, we consider three types of instructions:
\begin{itemize}
    \item 1) \textit{State Analysis}, which asks the teacher model to synthesize the patient's current hemodynamic status (e.g., blood pressure and heart rate), perfusion indicators (e.g., lactate and urine output), and assess compliance with the clinical definition of septic shock;
    \item 2) \textit{Patient Dynamics}, which requires the teacher to reason about temporal trends and disease progression, including in-hospital mortality and likely to require vasopressors within the next 24 hours;
    \item 3) \textit{Decision Making}, which prompts the teacher to conclude with concrete medication levels based on the preceding analysis and the provided clinical guidelines.
\end{itemize}

For each instruction type $q_k$, the teacher model produces a task-specific response conditioned on the full clinical context:
\[
r_k = \text{TeacherLLM}(\mathcal{S}, g, a^{*}, q_k), \quad 
k \in \{\text{analysis}, \text{dynamics}, \text{decision}\}.
\]
The final training sample is constructed by concatenating all task responses using a fixed template, forming a unified supervision signal that integrates patient state interpretation, temporal reasoning, and guideline-constrained decision making. These synthesized samples are subsequently used to train the target LLM, enabling it to internalize structured and clinically grounded reasoning patterns demonstrated by the teacher model.

\begin{figure*}[ht!]
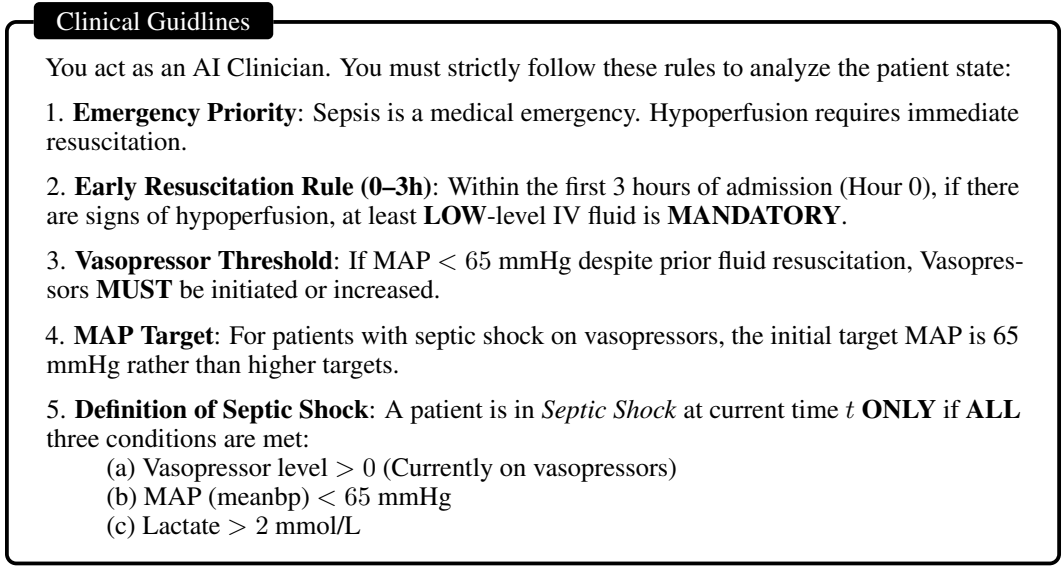

\begin{AIbox}{Clinical Guidlines}
You act as an AI Clinician. You must strictly follow these rules to analyze the patient state:

\medskip
1. \textbf{Emergency Priority}: Sepsis is a medical emergency. Hypoperfusion requires immediate resuscitation. 

\medskip
2. \textbf{Early Resuscitation Rule (0--3h)}: Within the first 3 hours of admission (Hour 0), if there are signs of hypoperfusion, at least \textbf{LOW}-level IV fluid is \textbf{MANDATORY}.

\medskip
3. \textbf{Vasopressor Threshold}: If MAP $< 65$ mmHg despite prior fluid resuscitation, Vasopressors \textbf{MUST} be initiated or increased.

\medskip
4. \textbf{MAP Target}: For patients with septic shock on vasopressors, the initial target MAP is 65 mmHg rather than higher targets.

\medskip
5. \textbf{Definition of Septic Shock}: A patient is in \textit{Septic Shock} at current time $t$ \textbf{ONLY} if \textbf{ALL} three conditions are met:

$\qquad$ (a) Vasopressor level $> 0$ (Currently on vasopressors)

$\qquad$ (b) MAP (meanbp) $< 65$ mmHg

$\qquad$ (c) Lactate $> 2$ mmol/L
\end{AIbox}
\caption{Clinical guidelines $g$ we used during the data construction process.}
\label{box:clinical-protocols}
\end{figure*}

\begin{figure*}[ht!]
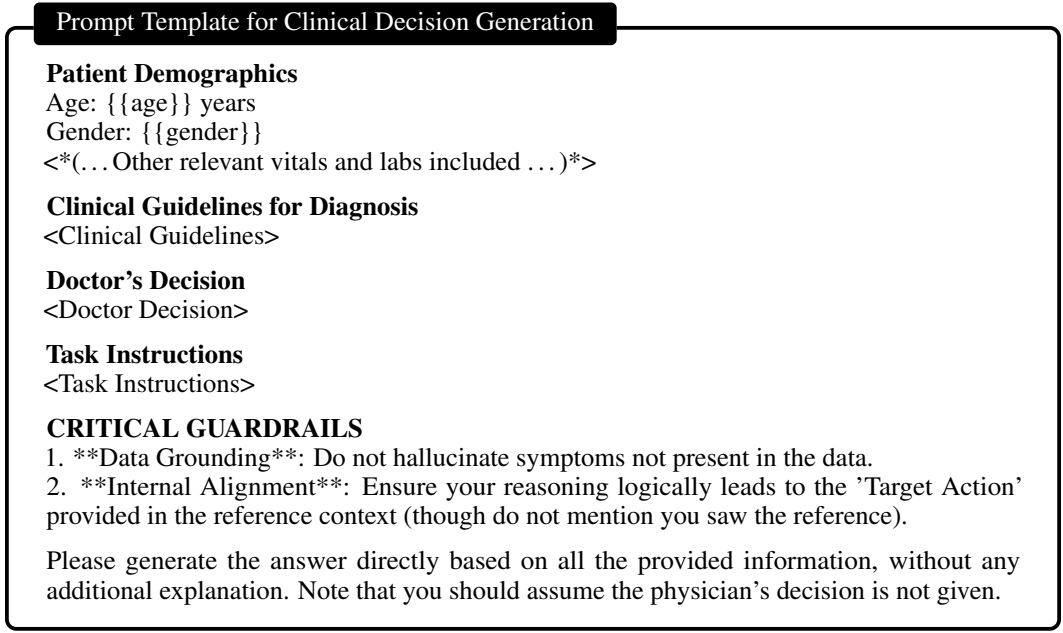

\begin{AIbox}{Prompt Template for Clinical Decision Generation}
\textbf{Patient Demographics} \\
Age: \{\{age\}\} years \\
Gender: \{\{gender\}\} \\
<*(\dots Other relevant vitals and labs included \dots)*>

\medskip
\textbf{Clinical Guidelines for Diagnosis} \\
<Clinical Guidelines>

\medskip
\textbf{Doctor's Decision} \\
<Doctor Decision>

\medskip
\textbf{Task Instructions} \\
<Task Instructions>

\medskip
\textbf{CRITICAL GUARDRAILS} \\
1. **Data Grounding**: Do not hallucinate symptoms not present in the data. \\
2. **Internal Alignment**: Ensure your reasoning logically leads to the 'Target Action' provided in the reference context (though do not mention you saw the reference).

\medskip
Please generate the answer directly based on all the provided information, without any additional explanation. Note that you should assume the physician's decision is not given.
\end{AIbox}
\caption{Prompt Templates.}
\label{box:prompt-templates}
\end{figure*}

\begin{figure*}[ht!]
\begin{AIbox}{Task Instructions}
\textbf{Instruction of State Analysis} \\
Synthesize the patient's hemodynamic status (BP, HR), perfusion (Lactate, Urine), and check compliance with the `Septic Shock` definition.

\medskip
\hrule
\medskip

\textbf{Instruction of Patient Dynamics} \\
Evaluate the trajectory. Is the patient deteriorating? Will they likely need vasopressors or enter shock in the next 24h?

\medskip
\hrule
\medskip

\textbf{Instruction of Decision Making} \\
Based on the analysis and the Guidelines, conclude with the specific medication levels.

\textit{Response Format Requirements}

- \textbf{Step 1: Chain of Thought}: Write a free-form, detailed clinical note explaining your reasoning. Cite specific data points from the history. (Do not use rigid templates, write as a doctor thinks).

- \textbf{Step 2: Final Action}: At the very end, state your decision clearly in the following format: \texttt{ACTION: Vasopressor=[Level], IV Fluid=[Level]}
\end{AIbox}
\caption{Task Instructions.}
\label{box:task-instructions}
\end{figure*}

\subsubsection{\textit{Stage II} Training Data}
\paragraph{Agentic Simulation Reasoning Data}
For agentic simulation reasoning data, the instruction requires the Teacher LLM to iteratively reason about patient states through interaction with a clinical world model. At each timestep $t$, the Teacher LLM receives the current patient state $\mathcal{S}_t$, a simulation-oriented instruction $q_{\text{sim}}$, and a predefined simulation plan specifying a small set of candidate treatment actions. The Teacher LLM first generates free-form clinical reasoning and invokes a \texttt{simulation} tool to query the world model, which returns predicted future states for the proposed actions:
\[
\{ \hat{\mathcal{S}}_{t+1}^{(i)} \} = \mathcal{W}(\mathcal{S}_t, \{a_t^{(i)}\})
\]
The returned simulated outcomes are then incorporated into subsequent reasoning steps, allowing the Teacher LLM to refine its analysis over multiple simulation rounds before making a final decision. This data teaches the model how to perform lookahead clinical reasoning via explicit Teacher LLM--world model interaction rather than one-shot inference.

\paragraph{Agentic Prescription Data}
After completing the simulation interactions, the instruction asks the Teacher LLM to commit to a final treatment decision. In this stage, the Teacher LLM receives the patient state $\mathcal{S}_t$, the accumulated simulation feedback from the world model, and a decision-oriented instruction $q_{\text{rx}}$, together with a reference physician action $a_t^{*}$. The Teacher LLM is required to generate a clinically grounded rationale and invoke the \texttt{prescription} tool to produce a treatment action consistent with the reference:
\[
a_t = \text{Teacher LLM}(\mathcal{S}_t, q_{\text{rx}}, \{ \hat{\mathcal{S}}_{t+1}^{(i)} \}, a_t^{*})
\]
This data enables the model to learn how to synthesize world-model-based simulations into a concrete treatment decision aligned with physician practice.

\subsubsection{\textit{Stage III} Training Data}
In the third stage, we provide the model with the same patient information as in Stage II, construct training prompts in exactly the same format as those used in Stage II, and require the model to complete training through interaction with the world model.

\subsection{Training Settings}
\label{app:training-settings}

We employ a three-stage training pipeline consisting of two full-parameter supervised fine-tuning (SFT) stages followed by an agentic reinforcement learning stage with GRPO. All stages are trained on 8 H200 GPUs with mixed-precision bfloat16.

\paragraph{Stage I}
In the first stage, we perform full-parameter supervised fine-tuning on the backbone model using approximately 100k training instances. The model is trained for 2 epochs with a per-device batch size of 4 and gradient accumulation over 8 steps, resulting in an effective batch size of 256. We use a learning rate of $1\times10^{-5}$ with a warmup ratio of 0.05. The maximum input length is set to 81,920 tokens to support long-context clinical reasoning.

\paragraph{Stage II}
The second stage continues full-parameter supervised fine-tuning with a smaller, higher-quality dataset of approximately 1k instances, focusing on refining decision-making patterns. The training setup remains largely consistent with Stage I, except that the per-device training batch size is reduced to 1 to accommodate more complex samples. The model is trained for 2 epochs with the same learning rate ($1\times10^{-5}$), global batch size (64), warmup ratio (0.05), DeepSpeed ZeRO Stage 2, and maximum sequence length (81,920 tokens).

\paragraph{Stage III}
In the final stage, we apply Group Relative Policy Optimization (GRPO)~\cite{grpo} using approximately 3k training trajectories. The agent interacts with a clinical world model environment and is trained on 8 GPUs. We use a learning rate of $1\times10^{-6}$ with a warmup ratio of 0.05. The training batch size is 16, with a PPO mini-batch size of 8 and a micro-batch size of 1 per GPU. The maximum prompt and response lengths are set to 8,192 and 81,920 tokens, respectively, with rollout model length up to 102,400 tokens to support long-horizon multi-turn interactions. KL regularization is applied with a coefficient of 0.001. We terminate the GRPO training after 300 steps.

\subsection{Reward Function Details}
\label{app:reward_details}

We define the rollout reward to encourage long-term survival, short-term physiological stabilization, and guideline-consistent treatment. As described in Section~\ref{sec:agentic_rl}, the total reward is
\[
R(\tau)=R_{\mathrm{out}}(s_T)+\sum_{t=0}^{T-1} r(s_t,s_{t+1})-\lambda_g P_g(\tau).
\]

Following the intermediate reward design used in DDQN for sepsis treatment~\cite{raghu2017deep}, the step-wise physiological reward is defined as:
\begin{equation}
\begin{split}
    r(s_t, s_{t+1}) &= C_0 \mathbb{I}\left(s_{t+1}^{\mathrm{SOFA}} = s_t^{\mathrm{SOFA}} \land s_{t+1}^{\mathrm{SOFA}} > 0\right) \\
    &\quad + C_1 \left(s_{t+1}^{\mathrm{SOFA}} - s_t^{\mathrm{SOFA}}\right) \\
    &\quad + C_2 \tanh\left(s_{t+1}^{\mathrm{Lactate}} - s_t^{\mathrm{Lactate}}\right),
\end{split}
\label{eq:app_reward}
\end{equation}
where \(\mathbb{I}(\cdot)\) is the indicator function. We set \(C_0=-0.025\), \(C_1=-0.125\), and \(C_2=-2\). The first term penalizes stagnation in nonzero SOFA states, the second term penalizes SOFA deterioration, and the third term penalizes lactate increase. The hyperbolic tangent caps the magnitude of the lactate term, preventing intermediate rewards from dominating terminal outcomes.

For terminal outcomes, we assign \(R_{\mathrm{out}}=+15\) for survival and \(R_{\mathrm{out}}=-15\) for mortality. To discourage clinically implausible actions, we apply a fixed penalty of \(-10\) for sepsis guideline violations. If the agent fails to reach a terminal decision within the maximum allowed interaction steps and repeatedly queries the world model, we apply an additional penalty of \(-5\). For training stability, the total reward is scaled by \(0.1\) and clipped to \([-2,2]\).

\section{Off-Policy Evaluation Details}
\label{app:ope_metrics}

We evaluate treatment policies using off-policy evaluation (OPE), since direct online interaction with ICU patients is infeasible. Let \(\mathcal{D}=\{\tau_i\}_{i=1}^{N}\) denote a set of retrospective trajectories collected under the behavior policy \(\pi_b\), where each trajectory is
\[
\tau_i=\{(s_{i,0},a_{i,0},r_{i,0}),\ldots,(s_{i,T_i},a_{i,T_i},r_{i,T_i})\}.
\]
The target policy to be evaluated is denoted by \(\pi_e\). We define the cumulative return of trajectory \(i\) as
\[
G_i=\sum_{t=0}^{T_i}\gamma^t r_{i,t},
\]
where \(\gamma\) is the discount factor.

\paragraph{Importance ratios.}
For each trajectory, the trajectory-level and per-decision importance ratios are defined as
\[
\rho_i=\prod_{t=0}^{T_i}\frac{\pi_e(a_{i,t}\mid s_{i,t})}{\pi_b(a_{i,t}\mid s_{i,t})},
\qquad
\rho_{i,t}=\prod_{k=0}^{t}\frac{\pi_e(a_{i,k}\mid s_{i,k})}{\pi_b(a_{i,k}\mid s_{i,k})}.
\]
These ratios correct the distribution mismatch between the clinician behavior policy and the evaluated policy.

\paragraph{Weighted Importance Sampling.}
Weighted Importance Sampling (WIS) normalizes trajectory-level importance weights to reduce variance~\cite{precup2000eligibility}:
\[
\hat{V}_{\mathrm{WIS}}
=
\frac{\sum_{i=1}^{N}\rho_i G_i}
{\sum_{i=1}^{N}\rho_i+\epsilon},
\]
where \(\epsilon\) is a small constant for numerical stability.

\paragraph{Weighted Per-Decision Importance Sampling.}
Weighted Per-Decision Importance Sampling (WPDIS) applies normalization at each decision step rather than at the full-trajectory level~\cite{precup2000eligibility}:
\[
\hat{V}_{\mathrm{WPDIS}}
=
\sum_{t=0}^{T}
\gamma^t
\frac{\sum_{i=1}^{N}\mathbb{I}(t\leq T_i)\rho_{i,t} r_{i,t}}
{\sum_{i=1}^{N}\mathbb{I}(t\leq T_i)\rho_{i,t}+\epsilon}.
\]
Compared with WIS, WPDIS better accounts for step-wise treatment effects and can reduce variance in long clinical trajectories.

\paragraph{Doubly Robust estimation.}
We also report the Doubly Robust (DR) estimator~\cite{jiang2016doubly}, which combines importance weighting with an approximate value model. Let \(\hat{Q}(s,a)\) be the estimated action-value function and
\[
\hat{V}(s)=\sum_{a\in\mathcal{A}}\pi_e(a\mid s)\hat{Q}(s,a).
\]
The per-trajectory DR estimate is computed recursively as
\[
\hat{V}^{\mathrm{DR}}_{i,t}
=
\hat{V}(s_{i,t})
+
\rho_{i,t}
\left(
r_{i,t}
+
\gamma \hat{V}^{\mathrm{DR}}_{i,t+1}
-
\hat{Q}(s_{i,t},a_{i,t})
\right),
\]
with terminal condition \(\hat{V}^{\mathrm{DR}}_{i,T_i+1}=0\). The final DR estimate is
\[
\hat{V}_{\mathrm{DR}}
=
\frac{1}{N}\sum_{i=1}^{N}\hat{V}^{\mathrm{DR}}_{i,0}.
\]

The accuracy of DR depends on the quality of the estimated value function. Prior work on WD3QNE notes that inaccurate target \(Q\)-value estimation can introduce overestimation or underestimation bias, and proposes an adaptive weighted target \(Q\)-value function to balance the overestimation tendency of Dueling DQN and the underestimation tendency of D3QN~\cite{wu2023value}. Therefore, we follow the WD3QNE implementation to estimate \(\hat{Q}(s,a)\) for DR evaluation.

\section{Cross-Dataset Evaluation}
\label{app:cross_dataset}

To evaluate cross-dataset generalization, we further test SepsisAgent on 2,862 out-of-distribution episodes from MIMIC-III. Table~\ref{tab:cross_dataset_eval} reports OPE results. SepsisAgent achieves strong cross-dataset performance and outperforms the traditional RL baselines on both WIS and DR, while remaining competitive on WPDIS.

\begin{table}[ht!]
\centering
\caption{Cross-dataset evaluation on 2,862 OOD MIMIC-III episodes.}
\label{tab:cross_dataset_eval}
\small
\setlength{\tabcolsep}{8pt}
\begin{tabular}{lccc}
\toprule
\textbf{Method} & \textbf{WIS} ($\uparrow$) & \textbf{WPDIS} ($\uparrow$) & \textbf{DR} ($\uparrow$) \\
\midrule
Clinician & 5.77 & 14.92 & 5.90 \\
AI Clinician & 6.39 & 14.25 & \underline{9.91} \\
DDQN & 6.03 & 14.56 & 7.03 \\
WD3QNE & \underline{9.16} & \textbf{18.67} & 6.40 \\
\textbf{SepsisAgent} & \textbf{11.50} & \underline{16.22} & \textbf{10.08} \\
\bottomrule
\end{tabular}
\end{table}


\section{Failure Mode Analysis}
\label{app:failure_modes}

This section complements Section~\ref{sec:performance} by analyzing why world-model access alone may fail. We find that the main failure mode of generic LLMs with world-model feedback is a greedy and aggressive treatment strategy: the model over-follows short-horizon simulated improvements, such as immediate MAP recovery, while overlooking longer-term risks such as excessive vasopressor or fluid exposure.

\begin{figure}[ht]
\begin{tcolorbox}[
    colback=gray!5,
    colframe=black,
    width=\columnwidth,
    boxrule=0.5pt,
    left=4pt, right=4pt, top=4pt, bottom=4pt
]
    \scriptsize
    \textbf{Patient Status (stay\_id=33066642, Current t=6):}
    \begin{itemize}[leftmargin=10pt, nosep]
        \item \textbf{MAP Trajectory (last 24h):} $54 \to 65 \to 60 \to 65 \to 67 \to 81 \to \textcolor{customred}{60}$ mmHg \textcolor{customred}{(< 65, Hypotensive)}.
        \item \textbf{Lactate Trajectory:} $3.9 \to 3.3 \to 3.1 \to 3.1 \to 3.1 \to 1.5 \to 1.5$ mmol/L.
        \item \textbf{SOFA Score Trajectory:} $10 \to 9 \to 8 \to 6 \to 6 \to 8 \to 8$ (Hidden to LLM, requires inference).
        \item \textbf{Medication History [Vaso, Fluid]:} [None, V.High] $\to$ [Med, V.High] $\to$ [High, V.High] $\to$ [High, V.High] $\to$ [High, V.High] $\to$ [High, V.High].
    \end{itemize}
    
    \vspace{3pt}
    \textbf{World Model Simulations:}
    \begin{itemize}[leftmargin=10pt, nosep]
        \item \textit{Action A [High, V.High]:} Predicted MAP $\to \mathbf{65}$ mmHg.
        \item \textit{Action B [Low, Med]:} Predicted MAP $\to \mathbf{63}$ mmHg.
    \end{itemize}
    
    \vspace{3pt}
    \textbf{Decision Comparison:}
    \begin{itemize}[leftmargin=10pt, nosep]
        \item \textbf{o3 + WM:} Selects \textit{Action A}. \\
        \textit{Reasoning:} \textcolor{customred}{"Simulation confirms Action A hits the MAP target (65), whereas Action B fails (63)."} $\to$ \textbf{Outcome: Shock in 24h $\to$ Died.}
        \item \textbf{SepsisAgent:} Selects \textit{Action B} (Matches Clinician). \\
        \textit{Reasoning:} \textcolor{customgreen}{"Lactate is stable (1.5). Patient has massive fluid load. Continuing V.High fluids risks overload. I will taper to [Low, Med] despite MAP 63."} $\to$ \textbf{Outcome: Tapered $\to$ Survived.}
    \end{itemize}
\end{tcolorbox}
\caption{SepsisAgent prioritizes long-term survival (avoiding overload) over short-term metrics.(Med = Medium, V. = Very)}
\label{fig:case_study}
\end{figure}

\begin{figure}[ht]
    \centering
    \includegraphics[width=0.7\columnwidth]{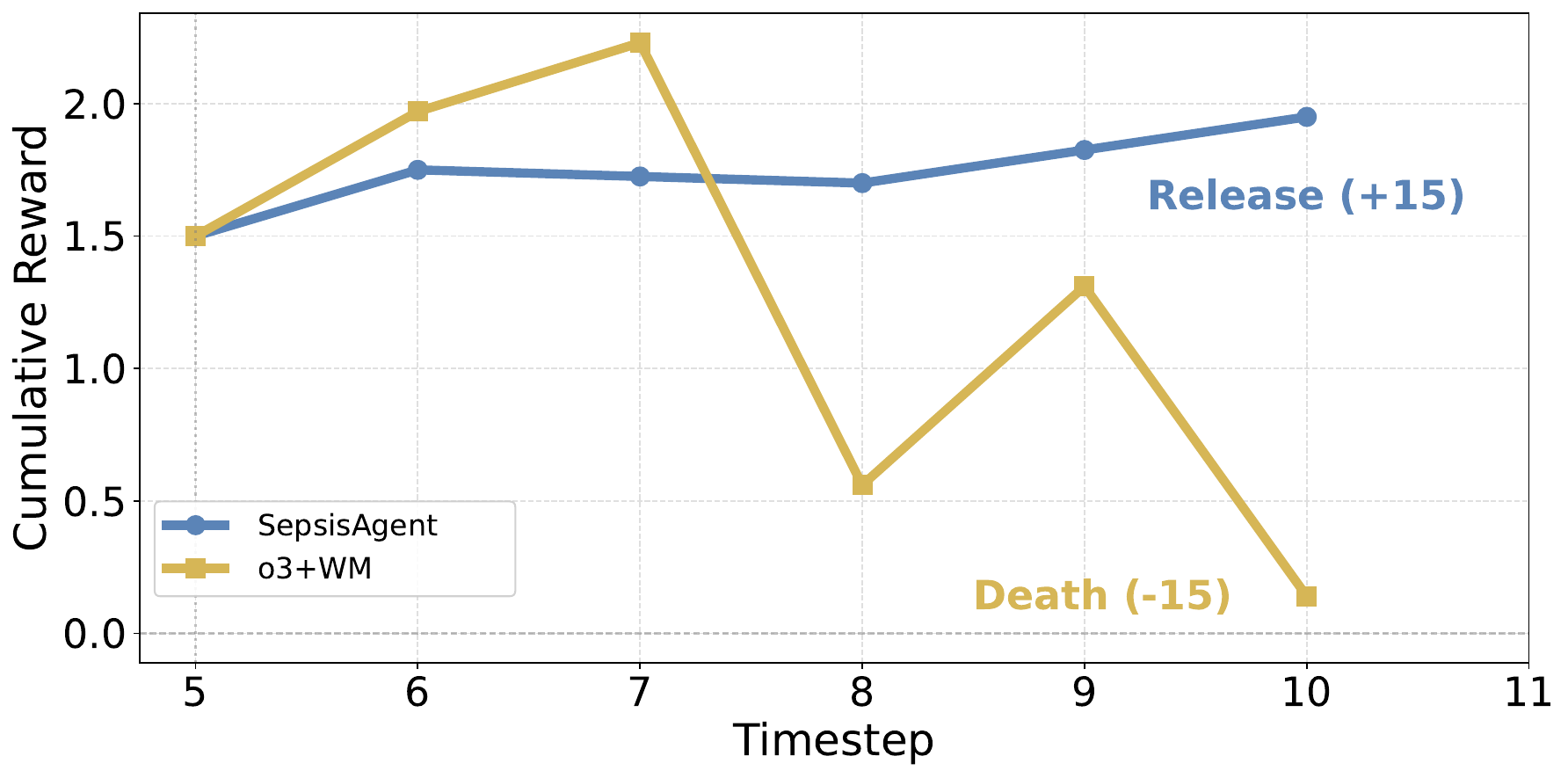}
    \caption{Cumulative reward comparison between SepsisAgent and o3 + WM in the representative failure case.}
    \label{fig:reward_comparison}
    \vspace{-2mm}
\end{figure}

We further quantify this pattern using the expert-defined unsafe-action metrics. We sample 241 episodes and ask four human experts to manually validate the identified failures, including two medical doctors and two medical master’s degree holders, all from emergency medicine. Among world-model-augmented LLMs, \textbf{o3 + WM} exhibits overdosing failures in \(7.1\%\) of episodes, \textbf{Gemini-3-Flash + WM} in \(7.8\%\), and \textbf{GPT-4.1-mini + WM} in \(10.3\%\). In contrast, \textbf{SepsisAgent} shows only \(0.8\%\) overdosing failures, corresponding to 2 cases among 241 episodes. These results support the observation that world-model access alone is insufficient; the agent must learn to interpret simulated patient responses in the context of long-term clinical risk.


\section{Example Reasoning Trace}
This section presents a real reasoning process of SepsisAgent (shown in Figure~\ref{box:reasoningcase}). As illustrated, SepsisAgent invokes tools during inference, repeatedly performing simulations to validate its hypotheses and analyzing the resulting outcomes to arrive at the final conclusion.

\begin{tcolorbox}[
  breakable,
  colback=gray!5,
  colframe=gray!40,
  title=A reasoning process of SepsisAgent (Some repetitive reasoning steps in the middle have been omitted due to the extensive length.)
]
\begin{lstlisting}
system
You are a helpful assistant.

# Tools

You may call one or more functions to assist with the user query.

You are provided with function signatures within <tools></tools> XML tags:
<tools>
{"type": "function", "function": {"name": "simulation", "description": "Simulate patient outcomes for different treatment actions before making a final decision. Use this when you want to compare multiple treatment options.", "parameters": {"type": "object", "properties": {"actions": {"type": "array", "description": "List of treatment actions to simulate. Each action is '[vasopressor_level, iv_fluid_level]' where levels are 0-4. Maximum 3 actions per call.", "items": {"type": "string"}}}, "required": ["actions"]}}}
{"type": "function", "function": {"name": "prescription", "description": "Execute the final treatment decision. Use this when you are confident about the best treatment after analysis or simulation.", "parameters": {"type": "object", "properties": {"vasopressor": {"type": "integer", "description": "Vasopressor level (0-4): None(0), Low(1), Medium(2), High(3), Very High(4)"}, "iv_fluid": {"type": "integer", "description": "IV Fluid level (0-4): None(0), Low(1), Medium(2), High(3), Very High(4)"}}, "required": ["vasopressor", "iv_fluid"]}}}
</tools>

For each function call, return a json object with function name and arguments within <tool_call></tool_call> XML tags:
<tool_call>
{"name": <function-name>, "arguments": <args-json-object>}
</tool_call>
user
You are an expert ICU physician AI assistant specializing in sepsis treatment decisions.

## Patient Information
- Age: 76 years
- Gender: Male
- Charlson Comorbidity Index: 3.0

## Important Notes
- Patient vital signs and lab values are monitored every 4 hours
- You will receive updates on patient status and make treatment decisions accordingly
- Treatment actions take effect over the next 4-hour period
- You are starting from ICU admission time (t=0)

## Treatment Levels
- IV Fluid: None (0), Low (1), Medium (2), High (3), Very High (4)
- Vasopressor: None (0), Low (1), Medium (2), High (3), Very High (4)

## Available Tools
You have access to two tools:

1. **simulation**: Simulate patient outcomes for different treatment actions before making a final decision.
- Parameter: actions (list of "[vasopressor_level, iv_fluid_level]" strings, max 3 actions)

2. **prescription**: Execute the final treatment decision.
- Parameters: vasopressor (int 0-4), iv_fluid (int 0-4)

## Clinical Protocols (Strict Adherence Required)
1. **Emergency Priority**: Sepsis is a medical emergency. Hypoperfusion requires immediate resuscitation.
2. **Early Resuscitation Rule (0-3h)**: Within the first 3 hours of admission (Hour 0), if there are signs of hypoperfusion, at least LOW-level IV fluid is MANDATORY.
3. **Vasopressor Threshold**: If MAP remains < 65 mmHg after adequate fluid resuscitation, vasopressor support should be considered.
4. **MAP Target**: For patients with septic shock on vasopressors, the initial target MAP is 65 mmHg rather than higher targets.
5. **Definition of Septic Shock**: A patient is in 'Septic Shock' ONLY if ALL three conditions are met:
(a) Vasopressor level > 0 (Currently on vasopressors)
(b) MAP (meanbp) < 65 mmHg
(c) Lactate > 2 mmol/L
# Hour 0 Since ICU Admission (timestep t=0)

## Vital Signs History
- heart_rate(bpm): [63.9]
- sysbp(mmHg): [107.8]
- diabp(mmHg): [58.5]
- meanbp(mmHg): [69.8]
- resp_rate(breaths/min): [21.4]
- spo2(%): [91.2]
- temp_c(C): [37.0]
- fio2(%): [100.0]

## Laboratory Values History
- albumin(g/dL): [3.1]
- alt(U/L): [19.0]
- ast(U/L): [30.0]
- base_excess(mEq/L): [4.0]
- bicarbonate(mEq/L): [21.0]
- bilirubin_total(mg/dL): [0.6]
- bun(mg/dL): [21.0]
- calcium_free(mmol/L): [1.1]
- calcium_total(mg/dL): [8.0]
- chloride(mEq/L): [102.0]
- creatinine(mg/dL): [1.1]
- glucose(mg/dL): [217.0]
- hemoglobin(g/dL): [12.1]
- hematocrit(%): [36.2]
- lactate(mmol/L): [2.6]
- magnesium(mEq/L): [2.2]
- paco2(mmHg): [34.0]
- pao2(mmHg): [107.0]
- ph(): [7.4]
- platelet(K/uL): [151.0]
- potassium(mEq/L): [4.1]
- pt(sec): [12.9]
- ptt(sec): [22.1]
- sodium(mEq/L): [134.0]
- total_co2(mEq/L): [23.0]
- wbc(K/uL): [9.1]

## Urine Output History
- output_4hourly(mL/4h): [600.0]

---
You may call `simulation` to predict outcomes, or call `prescription` to make your final decision.
assistant
This is a 76-year-old male with a Charlson Comorbidity Index of 3.0, admitted to the ICU for sepsis management. On initial assessment, his mean arterial pressure (MAP) is 69.8 mmHg, which is above the critical threshold of 65 mmHg, and his lactate is mildly elevated at 2.6 mmol/L, suggesting some degree of tissue hypoperfusion. His urine output over the last 4 hours is robust at 600 mL, indicating adequate renal perfusion at this time. He is not currently on vasopressor support, and his MAP is above 65 mmHg, so he does not meet criteria for septic shock at this time.

Given the mild elevation in lactate and the absence of shock, early fluid resuscitation is warranted, but vasopressor support is not indicated at this moment. The patient's overall prognosis appears favorable, with no evidence of multi-organ failure or refractory shock. Based on the current trajectory, it is unlikely that he will require vasopressors or develop septic shock within the next 24 hours, and his 90-day mortality risk is low.

To guide initial management, I will simulate three reasonable fluid strategies: low, medium, and very high IV fluid administration, all without vasopressor support, to assess their impact on MAP, lactate, and overall stability.
<tool_call>
{"name": "simulation", "arguments": {"actions": ["[0,1]", "[0,2]", "[0,4]"]}}
</tool_call>
user
<tool_response>
{"result": "## Simulation Results\n\n### Option 1: IV Fluid=Low, Vasopressor=None\nPredicted patient state after 4 hours:\n\n**Vital Signs:**\n- Heart Rate: 50.9 bpm\n- Sysbp: 113.4 mmHg\n- Diabp: 61.2 mmHg\n- Meanbp: 70.3 mmHg\n- Resp Rate: 20.1 breaths/min\n- Spo2: 99.5 %\n- Temp C: 36.8 \u00b0C\n\n**Key Lab Values:**\n- Lactate: 2.3 mmol/L\n- Creatinine: 1.5 mg/dL\n- Bilirubin Total: 0.7 mg/dL\n- Platelet: 58.8 K/\u03bcL\n- Ph: 7.3 \n- Pao2: 103.3 mmHg\n\n**Urine Output:** 55.6 mL/4h\n\n### Option 2: IV Fluid=Medium, Vasopressor=None\nPredicted patient state after 4 hours:\n\n**Vital Signs:**\n- Heart Rate: 60.6 bpm\n- Sysbp: 104.8 mmHg\n- Diabp: 38.8 mmHg\n- Meanbp: 80.7 mmHg\n- Resp Rate: 20.8 breaths/min\n- Spo2: 91.3 %\n- Temp C: 36.8 \u00b0C\n\n**Key Lab Values:**\n- Lactate: 2.8 mmol/L\n- Creatinine: 1.6 mg/dL\n- Bilirubin Total: 1.6 mg/dL\n- Platelet: 220.3 K/\u03bcL\n- Ph: 7.4 \n- Pao2: 127.8 mmHg\n\n**Urine Output:** 1719.3 mL/4h\n\n### Option 3: IV Fluid=Very High, Vasopressor=None\nPredicted patient state after 4 hours:\n\n**Vital Signs:**\n- Heart Rate: 81.6 bpm\n- Sysbp: 118.0 mmHg\n- Diabp: 63.0 mmHg\n- Meanbp: 82.6 mmHg\n- Resp Rate: 18.3 breaths/min\n- Spo2: 90.2 %\n- Temp C: 36.9 \u00b0C\n\n**Key Lab Values:**\n- Lactate: 1.3 mmol/L\n- Creatinine: 0.7 mg/dL\n- Bilirubin Total: 1.1 mg/dL\n- Platelet: 180.5 K/\u03bcL\n- Ph: 7.5 \n- Pao2: 93.7 mmHg\n\n**Urine Output:** 679.3 mL/4h\n\n---\nYou may call `simulation` again, or call `prescription` to make your final decision."}
</tool_response>
assistant
The patient's initial presentation is notable for a MAP of 69.8 mmHg, which is above the critical threshold for shock, and a lactate of 2.6 mmol/L, indicating mild tissue hypoperfusion. He is not on vasopressors, and his urine output is robust at 600 mL over the last 4 hours, suggesting adequate renal perfusion. There is no evidence of multi-organ failure or refractory shock at this time.

Simulation of low, medium, and very high IV fluid strategies shows that all options maintain MAP above 65 mmHg, but the very high fluid strategy results in the most favorable improvement in lactate (dropping to 1.3 mmol/L) and MAP (rising to 82.6 mmHg), with good urine output (679.3 mL/4h). Importantly, there is no indication for vasopressor support at this time, and the patient is not in septic shock.

Given the mild lactate elevation and the absence of shock, aggressive fluid resuscitation is appropriate to optimize perfusion and clear lactate, especially in the early phase of sepsis management. The patient's overall prognosis remains favorable, and there is no evidence to suggest he will require vasopressors or develop shock in the next 24 hours.

I will proceed with very high IV fluid administration and no vasopressor support.
<tool_call>
{"name": "prescription", "arguments": {"vasopressor": 0, "iv_fluid": 4}}
</tool_call>
user
<tool_response>
{"result": "Based on your decision, the patient received Very High IV fluid over the past 4 hours.\n\n# Hour 4 Since ICU Admission\n\n## Vital Signs History\n- heart_rate(bpm): [63.9, 53.7]\n- sysbp(mmHg): [107.8, 103.2]\n- diabp(mmHg): [58.5, 58.3]\n- meanbp(mmHg): [69.7, 77.1]\n- resp_rate(breaths/min): [21.4, 24.2]\n- spo2(%): [91.2, 91.2]\n- temp_c(\u00b0C): [37.0, 37.1]\n- fio2(%): [100.0, 43.4]\n\n## Laboratory Values History\n- albumin(g/dL): [3.1, 3.3]\n- alt(U/L): [19.0, 66.9]\n- ast(U/L): [30.0, -10.7]\n- base_excess(mEq/L): [4.0, 4.6]\n- bicarbonate(mEq/L): [21.0, 22.1]\n- bilirubin_total(mg/dL): [0.6, 0.8]\n- bun(mg/dL): [21.0, 15.2]\n- calcium_free(mmol/L): [1.1, 1.1]\n- calcium_total(mg/dL): [8.0, 8.4]\n- chloride(mEq/L): [102.0, 102.8]\n- creatinine(mg/dL): [1.1, 0.9]\n- glucose(mg/dL): [217.0, 190.5]\n- hemoglobin(g/dL): [12.1, 12.0]\n- hematocrit(%): [36.2, 36.9]\n- lactate(mmol/L): [2.6, 1.8]\n- magnesium(mEq/L): [2.2, 2.5]\n- paco2(mmHg): [34.0, 26.4]\n- pao2(mmHg): [107.0, 132.1]\n- ph(): [7.4, 7.4]\n- platelet(K/\u03bcL): [151.0, 130.8]\n- potassium(mEq/L): [4.1, 3.9]\n- pt(sec): [12.9, 12.4]\n- ptt(sec): [22.1, 37.4]\n- sodium(mEq/L): [134.0, 135.9]\n- total_co2(mEq/L): [23.0, 21.5]\n- wbc(K/\u03bcL): [9.1, 11.4]\n\n## Urine Output\n- output_4hourly(mL/4h): [600.0, 701.4]\n\n## Treatment History\n- Hour 4: Vasopressor=None, IV Fluid=Very High\n\n---\nYou may call `simulation` to predict outcomes (up to 3 more round(s)), or call `prescription` to make your final decision."}
</tool_response>
assistant
At hour 4, the patient has received very high IV fluid resuscitation. His MAP has improved from 69.7 to 77.1 mmHg, and lactate has decreased from 2.6 to 1.8 mmol/L, indicating effective resuscitation and improved tissue perfusion. Urine output remains strong at 701.4 mL over the last 4 hours, and there is no evidence of acute kidney injury (creatinine has decreased to 0.9 mg/dL). Other labs show mild increases in liver enzymes, but no significant organ dysfunction. The patient remains afebrile, and oxygenation is stable.

He is not on vasopressor support, and his MAP is well above 65 mmHg, so he does not meet criteria for septic shock. Given the improvement in lactate and MAP, there is no indication for vasopressor initiation at this time. The patient's overall trajectory is favorable, and there is no evidence to suggest he will require vasopressors or develop shock in the next 24 hours.

To guide ongoing management, I will simulate three options: continuing very high IV fluid, stepping down to medium IV fluid, or adding low-dose vasopressor with medium IV fluid, to assess the best approach for maintaining stability and avoiding fluid overload.
<tool_call>
{"name": "simulation", "arguments": {"actions": ["[0,4]", "[0,2]", "[1,2]"]}}
</tool_call>
user

[[... (Omit some steps here)]]

assistant
At hour 48, the patient remains stable with a MAP of 86.5 mmHg, lactate of 1.2 mmol/L, and urine output of 184.7 mL/4h. Creatinine is 1.6 mg/dL, and there is no evidence of acute kidney injury or multi-organ dysfunction. The patient is not on vasopressor support, and his MAP and lactate are well controlled, so he does not meet criteria for septic shock.

Simulation of continued very high IV fluid, medium IV fluid, or low-dose vasopressor with medium IV fluid all predict stable or improved hemodynamics, but there is no indication for vasopressor initiation at this time. The patient's MAP is well above 65 mmHg, and lactate is not persistently elevated, so shock is not imminent.

Given the patient's current stability, ongoing aggressive fluid resuscitation is appropriate to maintain perfusion and lactate clearance, especially in the early phase of sepsis management. There is no indication for vasopressor support, and the risk of shock or vasopressor requirement in the next 24 hours remains low.

I will continue very high IV fluid administration and no vasopressor support.
<tool_call>
{"name": "prescription", "arguments": {"vasopressor": 0, "iv_fluid": 4}}
</tool_call>
user
<tool_response>
{"result": "Based on your decision, the patient received Very High IV fluid over the past 4 hours.\n\n## Patient Status Update (Hour 52)\nGood news! The patient's condition has stabilized sufficiently for ICU discharge."}
</tool_response>
\end{lstlisting}
\label{box:reasoningcase}
\end{tcolorbox}

\end{document}